# A multi-task spatiotemporal deep neural network for predicting penetration depth and morphology in laser welding

Sen Li, Haichao Cui*, Chendong Shao, Yaqi Wang, Xinhua Tang

[1] *Shanghai Key Laboratory of Materials Laser Processing and Modification, School of Materials Science and Engineering, Shanghai Jiao Tong University, Shanghai 200240, PR China.*

**Abstract:** In laser penetration welding, the assessment of penetration state and weld seam morphology plays a crucial role in determining the weld quality. This paper presents a comprehensive introduction of the innovative muti-task deep learning model that has the capability to predict penetration state, depth, and weld seam morphology with high accuracy. The monitoring platform relies on weld pool images captured during the laser welding process using a complementary metal-oxide-semiconductor camera. The proposed model integrates spatiotemporal features extracted from top weld pool images along with welding parameters, establishing a deep learning framework based on convolutional neural networks and state space models for more efficient extraction and processing of spatial-temporal information. Furthermore, a reliable method for constructing the dataset is proposed to enhance both robustness and generalization capability of the developed model. Validation results on the test set demonstrate that prediction accuracy for penetration state can reach 99.35%, while prediction error for penetration depth is 1.79 millimeter, and accuracy of reconstructing the weld cross-section is 95.65%. This study provides new insights and methodologies for in-situ quality control strategies in laser penetration welding systems.

## 1. Introduction

| Nomenclature | |
|---|---|
| BPP | Beam Parameter Product |
| BPNN | Back Propagation Neural Network |
| CCD | Charge-Coupled Device |
| CMOS | Complementary Metal-Oxide-Semiconductor |
| CNN | Convolutional neural network |
| DNN | Deep neural networks |

* Corresponding author: haichaocui@sjtu.edu.cn

| | |
|---|---|
| EDM | Electrical Discharge Machining |
| FCN | Fully Convolutional Network |
| HD95 | 95th percentile Hausdorff Distance |
| LSTM | Long Short-Term Memory |
| MAE | Mean Absolute Error |
| MAU | Motion-Aware Unit |
| MFN | Memory Fusion Network |
| PWM | Pulse Width Modulation |
| RBFNN | Radial Basis Probabilistic Neural Network |
| RMSE | Root Mean Squared Error |
| ROC | Receiver Operating Characteristic |
| SVM | Support Vector Machine |
| TSL | Time series length |

In recent years, laser welding has been widely utilized in major equipment and manufacturing sectors such as aviation and aerospace, energy, shipbuilding, automotive, rail transportation, and engineering equipment (Cai et al., 2020). Compared to traditional welding methods (arc welding and resistance welding), laser welding offers advantages including reduced heat input, increased welding speed, and a greater depth-to-width ratio of the weld seam. However, achieving high-quality weld joints in laser welding requires careful selection of parameters such as laser power and welding speed (Squillace et al., 2012). Therefore, if reliable modeling of the entire laser penetration welding process can be achieved, it would enable real-time adjustment of the welding parameters to obtain fully penetrated superior weld joints.

During laser welding, a significant amount of diverse information is generated, including optical signals (You et al., 2016) and acoustic signals (Sebestova et al., 2012). Optical visual data have attracted considerable attention from researchers due to their ease of acquisition and rich informational content. Early research in this area focused on developing effective monitoring systems and identifying key visual characteristics. For instance, an affordable vision monitoring system has been developed that utilizes multiple optical filters to reduce intense radiation while enabling real-time capture of keyhole evolution during welding (Zhou et al., 2024). A 532 nm auxiliary light source was employed during fiber laser welding to capture clear boundary morphology of the weld pool (Luo and Shin, 2015a). With these improved imaging capabilities, researchers began to quantitatively link visual features to welding stability. A weld pool shadow segmentation algorithm was developed, which quantitatively characterizes the correlation between weld pool surface morphology and welding stability by defining and analyzing shadow parameters (Gao and Zhang, 2015). This provides novel approaches and insights for monitoring high-power laser welding penetration depth and morphology. During aluminum-steel lap welding, image processing has been utilized to statistically analyze the areas of the keyhole and the full penetration hole, linking them to various welding profiles (Kim and Ahn, 2012).

While the aforementioned studies established the value of visual monitoring, the analysis of weld pool images is complicated due to the extensive information they contain, posing a new challenge in effectively extracting and analyzing features automatically. To address this, researchers began applying traditional machine learning models. A coaxial monitoring system was employed to observe real-time morphological changes of keyholes during welding, utilizing an RBFNN-based predictive model fed with collected images to estimate post-weld defects (Luo and Shin, 2015b). The prediction of potential welding defects from top-captured weld pool images using BPNN and RBFNN was also investigated (Gao et al., 2014; Wan et al., 2017). By comparing the performance of BPNN, RBFNN, and multiple linear regression models, it was concluded that BPNN provides more accurate defect predictions. Additionally, a laser welding multi-sensor fusion monitoring system based on SVM was proposed (You et al., 2014). This system incorporates one photoelectric and two visual sensors, extracts feature through image processing techniques, and employs SVM for predicting welding quality, yielding favorable outcomes. These studies represent the initial attempts at data-driven analysis, but they primarily rely on classical machine learning models that require significant feature engineering and may not fully exploit the richness of the image data.

In recent years, with the rapid development of deep learning technologies, the emergence of CNNs has provided new perspectives and methods for addressing this issue (Hinton and Salakhutdinov, 2006). This advancement has led to increased popularity among researchers for integrating CNNs with welding techniques. A CNN with a small number of parameters was designed to achieve good performance in predicting porosity ≥100μm during the laser butt welding process of 6061 aluminum alloy (Zhang et al., 2020). A multi-task CNN model has been developed that utilizes weld pool images as input to simultaneously predict the weld penetration and width, demonstrating the model's effectiveness compared to other machine learning algorithms (Li et al., 2022). For the prediction of weld cross-sectional morphology, previous studies have predominantly focused on using BPNNs. These models typically use welding parameters as input to predict key geometric features such as depth of penetration and bead width. (Gihr et al., 2024; Le-Hong et al., 2023; Wu et al., 2024) However, there is a lack of studies that directly employ deep learning models to predict the weld cross-sectional morphology. Furthermore, existing research tends to predefine weld geometric feature parameters and use models to predict these parameters. This approach simplifies the problem by focusing on a limited number of geometric features but lacks generalization performance significantly. The geometric characteristics of fully penetrated welds differ entirely from those of non-penetrated welds during the laser penetration welding process, which cannot be adequately represented by a limited set of geometric feature parameters. Therefore, it is essential to utilize deep learning models directly for predicting the weld cross-sectional morphology, as this approach not only provides an intuitive observation of the weld morphology but also applies to various welding scenarios with different penetration conditions.

The laser welding process is highly intricate phenomena including weld pool and vapor flow, heat transfer, phase transformation, and the interaction between plasma and the laser beam. Consequently, the signals generated during laser welding exhibit distinct temporal characteristics, and relying solely on CCD imaging is insufficient to comprehensively depict the entire welding process (Brock et al., 2014). A CNN-LSTM model has been employed to forecast the weld penetration depth in GTAW welding processes (Yu et al., 2022). Experimental results demonstrated that compared to the CNN model, the CNN-LSTM model can swiftly track the dynamic changes in penetration depth, performing better on the validation set with an average error of only 0.3 mm from actual values. A wavelet packet transform analysis has also been conducted on images of vapor plumes in laser penetration welding, constructing a temporal feature-rich cubic time-frequency spectral dataset (Yan et al., 2024). Notably, the concept of multi-task spatiotemporal learning has also been effectively applied in other advanced joining processes. For instance, (Chen et al., 2021) developed an innovative multi-task learning method based on Gaussian processes to model the spatiotemporal progression of tool surfaces in ultrasonic metal welding, demonstrating significant improvements in prediction accuracy with limited data. These studies collectively underscore the importance of capturing spatiotemporal dynamics in manufacturing processes. Therefore, to comprehensively and accurately model the laser welding process, it is essential to deeply integrate multimodal information from various dimensions, such as top surface weld pool images and welding parameters.

Based on the aforementioned analysis, this paper investigates the apparent temporal and spatial characteristics observed in the top surface weld pool images and welding parameters during the laser penetration welding process. In comparison with conventional spatiotemporal prediction models, we propose a MAU (Chang et al., 2021) network-based weld pool image spatiotemporal feature extraction network, along with an MFN-Mamba network-based feature fusion network. This integrated network effectively combines the spatiotemporal features of welding parameters with those of the weld pool images. Moreover, based on these fully integrated spatiotemporal features, we establish networks for penetration depth prediction, and weld cross-sectional morphology reconstruction. For the weld cross-sectional morphology reconstruction network, a CNN-based decoder network is utilized to convert the spatiotemporal features of top surface weld pool images and welding parameters into images representing weld cross-sectional morphology. The workflow of our proposed model can be summarized in three steps: (a) Collecting top surface weld pool images and welding parameters under different laser welding conditions. (b) Enhancing raw data as necessary and training our deep learning model; (c) Validating the model using a test dataset and evaluating its performance. A 20kW spot-ring fiber laser was employed for this study. Various image processing techniques were applied to enhance the dataset used for model training purposes. After completing model training, we assessed its performance using a test dataset. We have sufficient grounds to believe that our model significantly distinguishes itself from other studies while making positive contributions to this research direction.

# 2. Experimental setup and method

## 2.1 Experimental setup

The experimental setup consists of three main components: the laser welding system, the robotic motion control system, and the image capture system. The laser welding system comprises a fiber laser device, a water chiller and a laser head. The movement of the laser welding head is controlled by a Yaskawa robot. Visual inspection follows Jonas's method (Näsström et al., 2020), which involves adjusting CMOS camera parameters to obtain clear, stable and distinct images of the weld pool. These images are then transmitted to a PC for storage. The output power of the fiber laser is regulated by the PC, while other parameters such as welding speed and defocus are managed by the robot.

In this study, we utilized the FEIBO-20000 model dot-ring fiber laser with maximum output power of 20 kW and wavelength of 1080 nm. It is focused through a lens with focal length of 165 mm. The BPP for the dot laser is $3.385\ mm{\cdot}mrad$ with focused spot diameter of $197.535\ \mu m$ and Rayleigh length of 2.88 mm. For the ring laser, the BPP is $15.377\ mm{\cdot}mrad$ with a focused spot diameter of $596.364\ \mu m$ and Rayleigh length of 26.25 mm. An image of the laser device is depicted in **Fig. 1(a)**.

As illustrated in **Fig. 1**, a CMOS camera was utilized during the experiment to dynamically monitor the weld pool from above, with a frame rate of 200 FPS. A narrowband filter and an auxiliary light source at a wavelength of 455 nm were utilized in conjunction with the CMOS camera. To ensure consistent light intensity and enhance image quality, a PWM wave generator was used to control both the CMOS camera and the auxiliary light source simultaneously. **Fig. 1(b)** illustrates the relative positions of the laser welding head, CMOS camera, 455 nm laser auxiliary light source, and the workpiece. **Fig. 2** depicts the hardware connection between the camera and the auxiliary light. A PC managed the PWM generator, producing two identical PWM signals: one for controlling the auxiliary light and another for triggering the CMOS camera. This synchronization ensured consistent illumination during image capture by the CMOS camera.

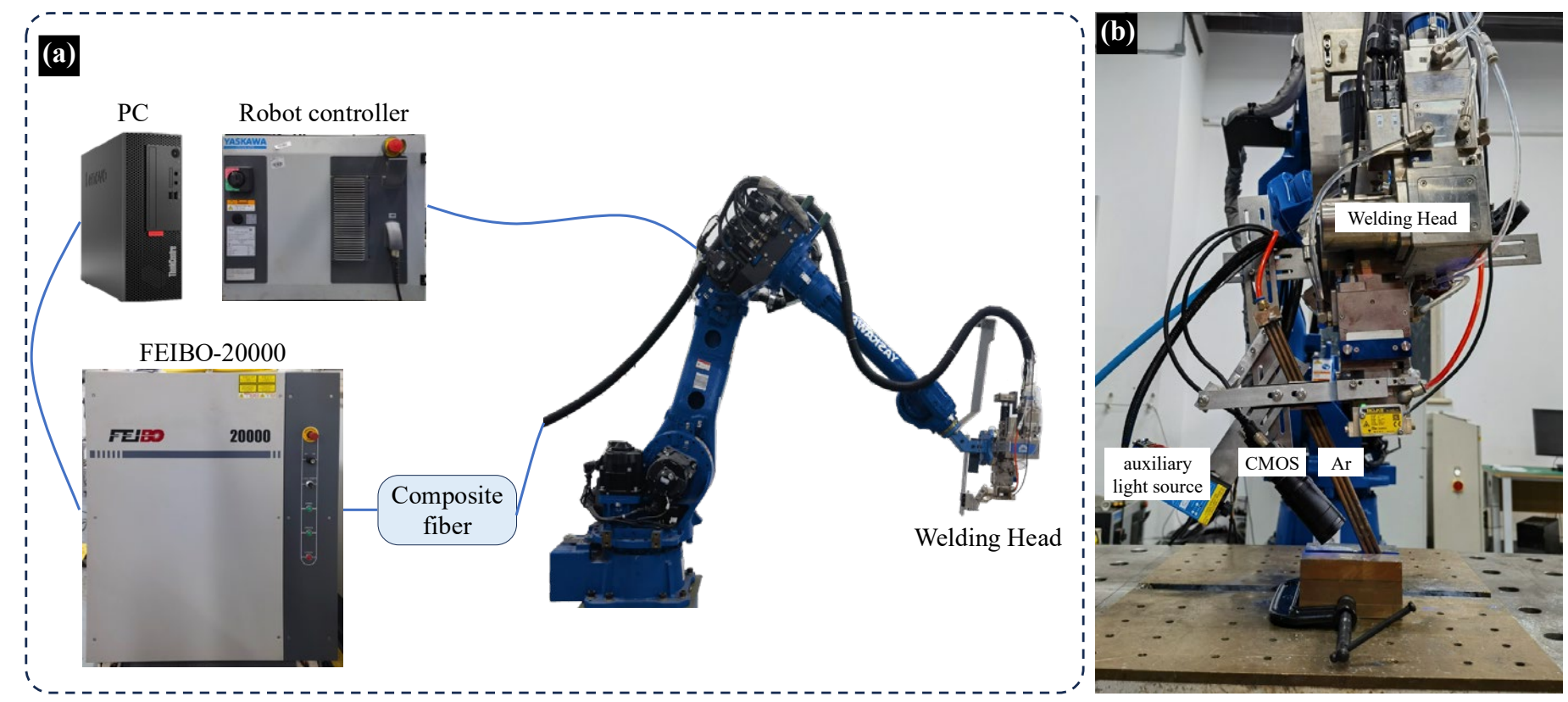

**Fig. 1** (a) Schematic of welding system; (b) Physical diagram of experimental setup.

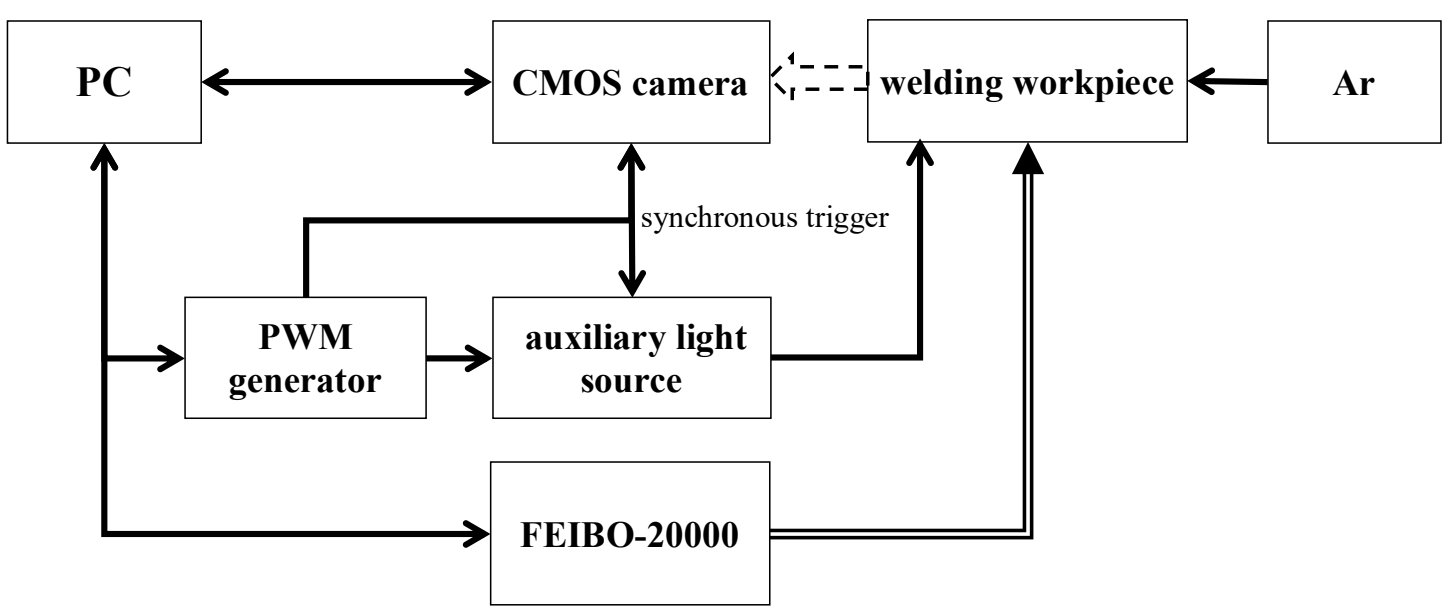

**Fig. 2** Schematic illustration of the connections between the CMOS camera, auxiliary light source, and FEIBO-20000

## 2.2 Experimental method

The specimens, made of COST-E material and measuring $100mm \times 40mm \times 50mm$ (length × width × height), exhibit blunt edges ranging from 8 mm to 14 mm as illustrated in **Fig. 3**.

This material is of significant value in critical industrial applications, such as advanced ultra-supercritical power plants, and has extremely stringent requirements for weld quality. Given these complexities, thick-section COST-E steel serves as an ideal benchmark to rigorously evaluate the performance and generalization capability of our model under demanding conditions. The chemical composition is provided in **Tab. 1**. Prior to welding, the surfaces of the specimens were meticulously polished cleaned with anhydrous alcohol to eliminate any oxides or other impurities that could potentially impact weld quality. Throughout the welding process, argon shielding is consistently applied as described in **Fig. 2** of the experimental setup. The shielding gas is delivered through three parallel copper tubes with outer diameters of 6 mm and inner diameters of 5 mm, ensuring a continuous flow rate of argon gas at 25 SLPM with a purity of 99.99%. Positioned at a distance of 5 mm from the workpiece surface and 3 mm from the laser beam, these tubes form an angle of approximately 30° with respect to the beam direction. The welding test was conducted following the scheme listed in **Tab. 2**, in which, $P_d$ is dot laser power, $P_r$ is ring laser power, $V$ is welding speed, $D_f$ is defocus, and $T$ is thickness of the welding workpiece. Additionally, it should be noted that both flat and horizontal welding position were employed for conducting experiments.

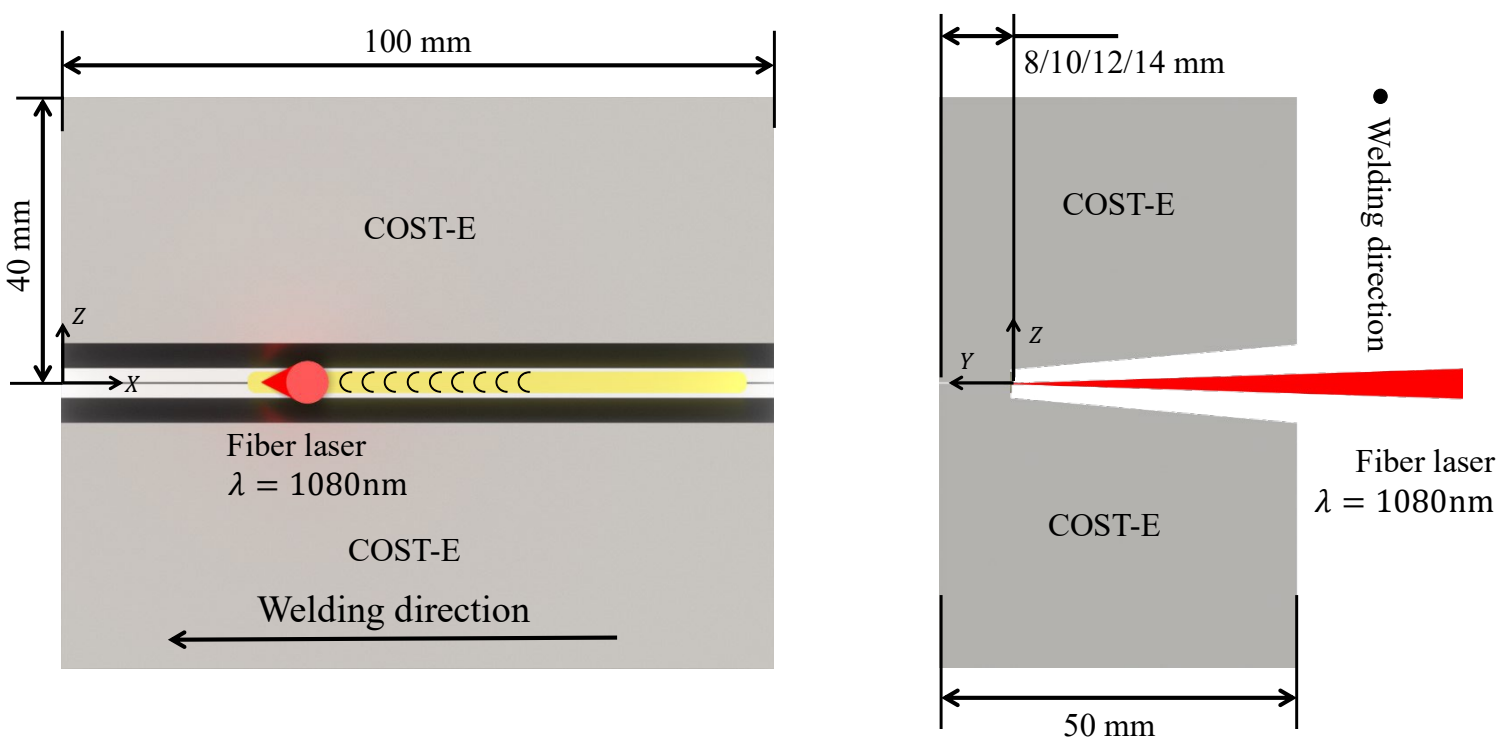

**Fig. 3** Schematics of the laser welding experiment.

**Tab. 1** Chemical composition of COST-E(wt.%) (Gianfrancesco, 2016).

| COST-E | C | Mn | Si | Ni | Cr | Mo | W | Nb | V | Fe |
|---|---|---|---|---|---|---|---|---|---|---|
| (wt.%) | 0.12 | 0.45 | 0.1 | 0.75 | 10.5 | 1.0 | 1.0 | 0.05 | 0.2 | Bal. |

**Tab. 2** Welding parameters.

| Test (No.) | $P_d(kW)$ | $P_r(kW)$ | $V(m/min)$ | $D_f(mm)$ | $T(mm)$ |
|---|---|---|---|---|---|
| 1 | 9.0 | 1.0 | 1.0 | 0.0 | 8.0 |
| 2 | 9.0 | 2.0 | 1.5 | -5.0 | 10.0 |
| 3 | 9.0 | 0.0 | 2.5 | -7.0 | 12.0 |
| 4 | 9.0 | 1.0 | 4.5 | 0.0 | 14.0 |
| 5 | 9.0 | 1.0 | 1.5 | -5.0 | 12.0 |
| 6 | 8.0 | 1.0 | 1.5 | -7.0 | 14.0 |
| 7 | 8.0 | 2.0 | 1.0 | -5.0 | 12.0 |
| 8 | 8.0 | 0.0 | 4.5 | 0.0 | 10.0 |

| Test (No.) | $P_d(kW)$ | $P_r(kW)$ | $V(m/min)$ | $D_f(mm)$ | $T(mm)$ |
|---|---|---|---|---|---|
| 9 | 8.0 | 0.0 | 2.5 | -5.0 | 8.0 |
| 10 | 7.0 | 1.0 | 2.5 | -5.0 | 10.0 |
| 11 | 7.0 | 2.0 | 4.5 | -7.0 | 8.0 |
| 12 | 7.0 | 0.0 | 1.0 | -5.0 | 14.0 |
| 13 | 7.0 | 2.0 | 1.5 | 0.0 | 12.0 |
| 14 | 9.0 | 1.0 | 4.5 | -5.0 | 12.0 |
| 15 | 8.0 | 2.0 | 2.5 | 0.0 | 14.0 |
| 16 | 9.0 | 1.0 | 1.5 | -5.0 | 14.0 |
| 17 | 8.0 | 0.0 | 1.5 | -7.0 | 8.0 |
| 18 | 7.0 | 1.0 | 1.0 | -7.0 | 10.0 |
| 19 | 9.0 | 1.0 | 1.0→1.5 | -7.0 | 14.0 |
| 20 | 8.0 | 2.0 | 1.5 | 0.0→-7.0 | 12.0 |
| 21 | 8.0 | 1.0 | 1.0 | 0.0 | 14.0→8.0→10.0 |
| 22 | 7.0 | 2.0 | 1.5 | -5.0 | 8.0→12.0→10.0 |

### 2.3 Data acquisition and definition

During the training process of deep learning models, a substantial volume of high-quality data is crucial as it directly influences the model's performance. Therefore, this section will provide detailed discussions on methods for data preparation and organization. Following the welding process described in Section 2.2, images of the top surface of weld pool were captured under different welding parameters, as shown in **Fig. 4**. The welding parameters corresponding to each image are labeled beneath each column. These images primarily consist of keyholes, welding seams, splashes, and reflections. The reflections occur when laser light from auxiliary light sources reflects off the sidewall of the weld pool. By utilizing a CMOS camera with external auxiliary lighting, we could effectively capture dynamic changes in the weld pool during the welding process and observe the morphology of the weld seam after solidification, thus demonstrating the effectiveness of this imaging approach. The captured images have a resolution of $640\,pixel \times 480\,pixel$. To reduce training time costs, the final image sequences included in the dataset are sampled at a frequency of 50FPS. Additionally, to enhance the robustness and generalization capability within our deep learning network, several image augmentation techniques were applied to these captured images, including random horizontal and vertical flips, translations along both axes randomly selected from horizontal or vertical directions respectively, as well as random color mapping transformations. Moreover, during laser deep penetration welding, the keyhole interface fluctuates due to high interfacial forces such as capillary and thermocapillary forces, and recoil pressure(Ai et al., 2018), resulting in slight variations in cross-sectional morphology within each welded joint. To enable the deep learning network to discern subtle shape differences caused by multiphysics phenomena, random weak shear transformations were also applied to these images.

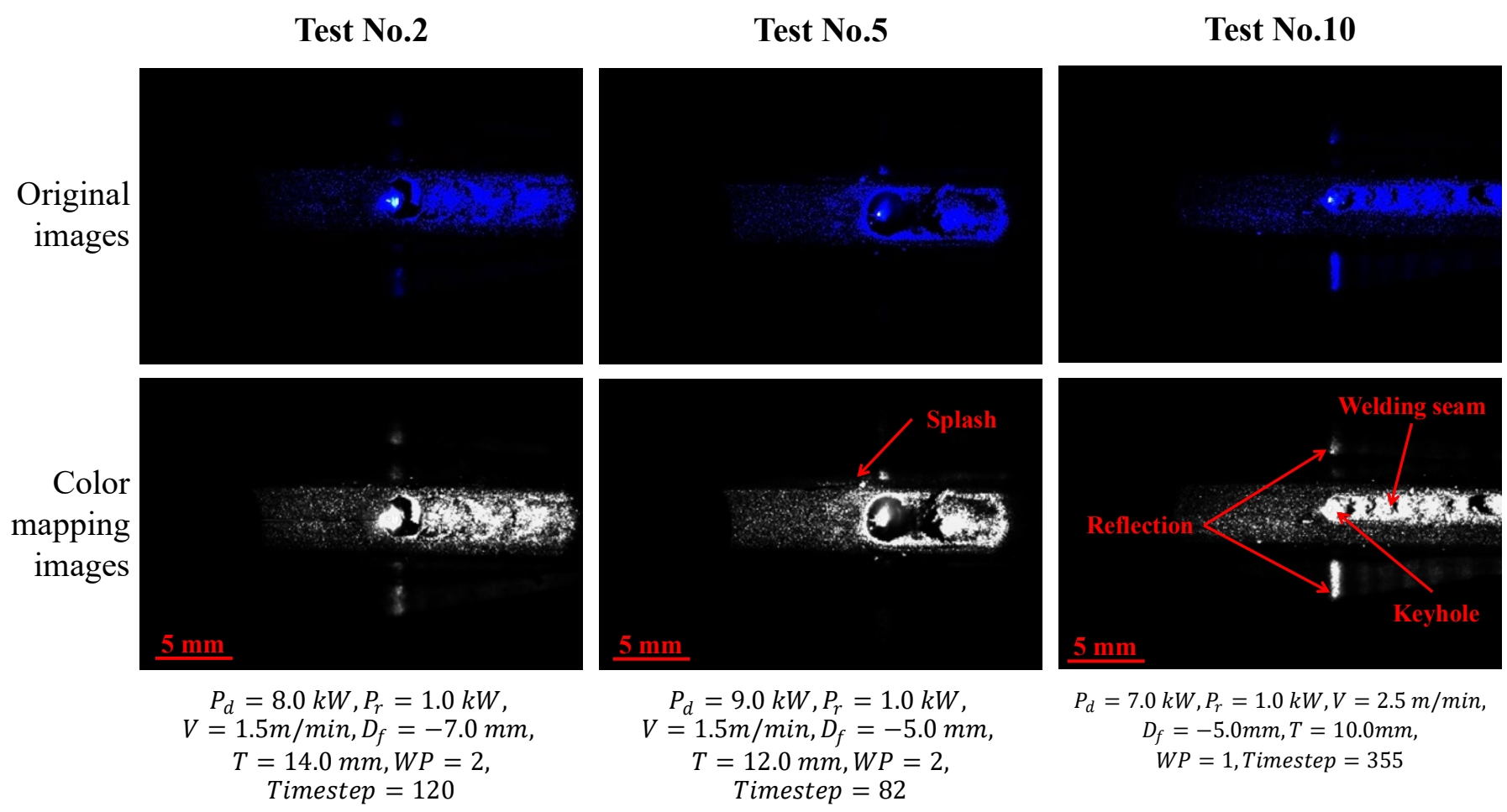

**Fig. 4** Images of the top surface of the weld pool during the welding process.

In the network proposed in this paper, the model for prediction requires the inclusion of welding parameters. This study identifies six critical welding parameters during the laser deep penetration welding process: $P_d$, $P_r$, $V$, $D_f$, $T$ as described above, and welding position $WP$, of which the flat position is denoted as 1 while the horizontal position is denoted as 2. The arrangement of these welding parameters in **Fig. 4** images is elaborated in **Tab. 3**. Following a single welding process, the size of the resulting vector containing welding parameters should be $Ts \times 6$ where $Ts$ represents the number of image samplings.

**Tab. 3** The arrangement methodology for the welding parameters within the dataset.

| | $P_d\ (kW)$ | $P_r (kW)$ | $V (m/min)$ | $D_f\ (mm)$ | $T (mm)$ | $WP$ |
|---|---|---|---|---|---|---|
| **Fig. 4(Test No.2)** | 8.0 | 1.0 | 1.5 | -7.0 | 14.0 | 2 |
| **Fig. 4(Test No.5)** | 9.0 | 1.0 | 1.5 | -5.0 | 12.0 | 1 |
| **Fig. 4(Test No.10)** | 7.0 | 1.0 | 2.5 | -5.0 | 10.0 | 2 |

Following the experimental procedure outlined in **Tab. 2**, the weld joints were initially cut along the weld seam using EDM. Subsequently, after undergoing polishing and etching process, the cross-sections of the welds were meticulously examined under a metallographic microscope. **Fig. 5** presents comprehensive longitudinal cross-sectional images of a weld obtained under typical welding parameters. The upper image visually depicts the macroscopic metallography of the weld joint, while the lower one showcases detailed image of the retained region within the welded joint. By conducting an analysis on pixel count derived from these longitudinal cross-sectional images, it becomes feasible to determine penetration depth at different time intervals during welding operations. Moreover, this analytical approach allows for corresponding assessment of penetration status at these positions.

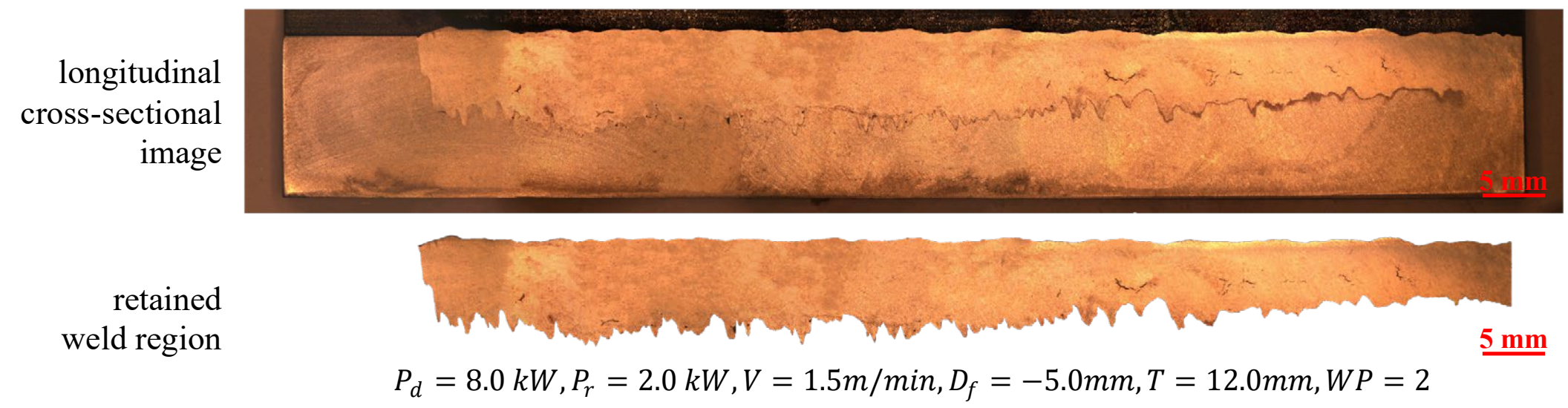

**Fig. 5** Images of the longitudinal cross-sections of a weld joint and the shape of the retained weld region.

After conducting the statistical analysis of penetration depth data, electrical discharge machining is utilized to cut perpendicularly along the weld seam. The required cross-section for each weld is dependent on the welding speed. In this study, a cross-sectional morphological image of the weld seam is generated every second. Once the cross-section is obtained, the metallographic processing method remains consistent with that of the longitudinal section. After capturing the metallographic images of the weld cross-section, they should be uniformly cropped to a size of $1400\,pixels \times 5600\,pixels$. Based on the magnification of the metallographic microscope, each millimeter corresponds to 350 pixels, with 1400 lateral pixels representing the upper width of the weld seam at 4 mm. The inclusion of 5600 vertical pixels ensures complete display even when dealing with a blunt thickness measuring up to 14 mm. Finally, all cross-sectional images are uniformly resized to dimensions of $256\,pixels \times 1024\,pixels$. A cross-sectional image of a weld seam is displayed in **Fig. 6**, in which the first row shows the cross-section images of the weld beads, the second row presents the annotated images, and the third row depicts the annotated images reduced by 50%, which are the final images included in the dataset. The welding parameters corresponding to each image are labeled beneath each column.

After the aforementioned processing, three datasets utilized in the multi-task network presented in this paper were generated: the penetration state dataset, the penetration depth dataset, and the weld cross-section dataset. Each dataset is divided into training, validation, and test sets in a 7:2:1 ratio, as shown in **Tab. 4**.

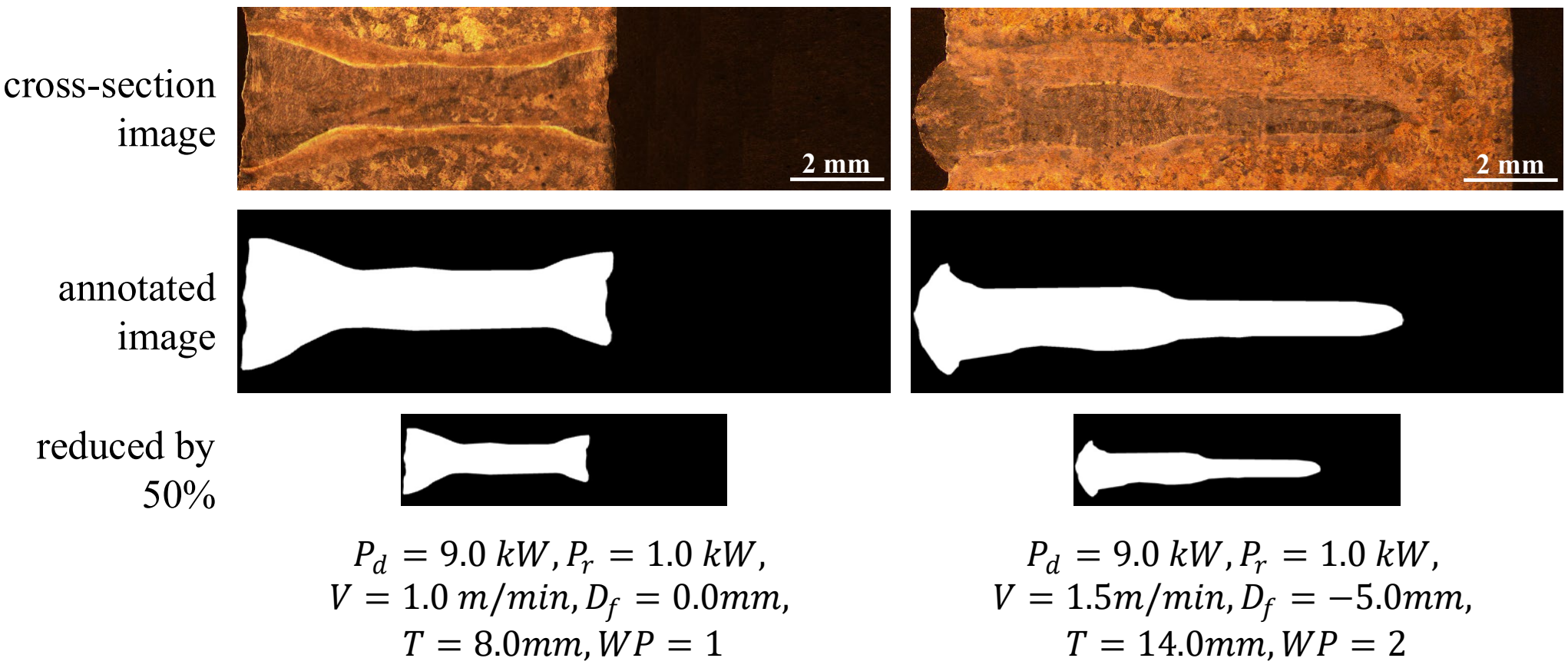

**Fig. 6** The cross-sectional image of the weld beads and its annotation.

**Tab. 4** Modeling dataset.

| | Penetration state dataset | | Penetration depth dataset | Weld cross-section morphology dataset |
|---|---|---|---|---|
| | Penetration | Non-Penetration | | |
| Train | 17595 | 17193 | 16154 | 388 |
| Valid | 5027 | 4912 | 4615 | 110 |
| Test | 2514 | 2457 | 2309 | 56 |
| Total samples | 25136 | 24562 | 23078 | 554 |

# 3. Proposed deep-learning model

### 3.1 Spatiotemporal feature extraction network

The welding process is highly intricate, as the current state of the weld pool is closely tied to its previous condition. As heat accumulates within the workpiece during welding, it leads to variations in both penetration state and depth. Consequently, welding is a continuous process that occurs over time and space. Based on this analysis, developing a deep learning network capable of effectively exploring and utilizing the temporal and spatial characteristics of welding signals is an essential prerequisite for achieving satisfactory results in subsequent tasks. This paper utilizes images captured from the top surface of the weld pool during welding, along with real-time welding parameters, as inputs to construct a spatiotemporal feature extraction network specifically designed for laser penetration welding. The performance of this network architecture has been compared with other alternatives.

In spatiotemporal sequence feature extraction, a classic baseline is the two-stage CNN-LSTM paradigm (Zhou et al., 2024). This approach first employs a CNN, such as the ResNet50 architecture (He et al., 2016) shown in **Fig. 7(a)**, to extract spatial features from individual image frames. These features are then fed into an LSTM network to model temporal dependencies. A similar process using a FCN (Shelhamer et al., 2017) is applied to the welding parameters, as depicted in **Fig. 7(b)**. However, this decoupled "spatial-then-temporal" strategy has a key limitation: it may lose crucial motion-related spatiotemporal information within the weld pool during the initial, frame-by-frame feature extraction. Furthermore, LSTMs are known to face challenges with long sequences, such as vanishing gradients and computational inefficiency.

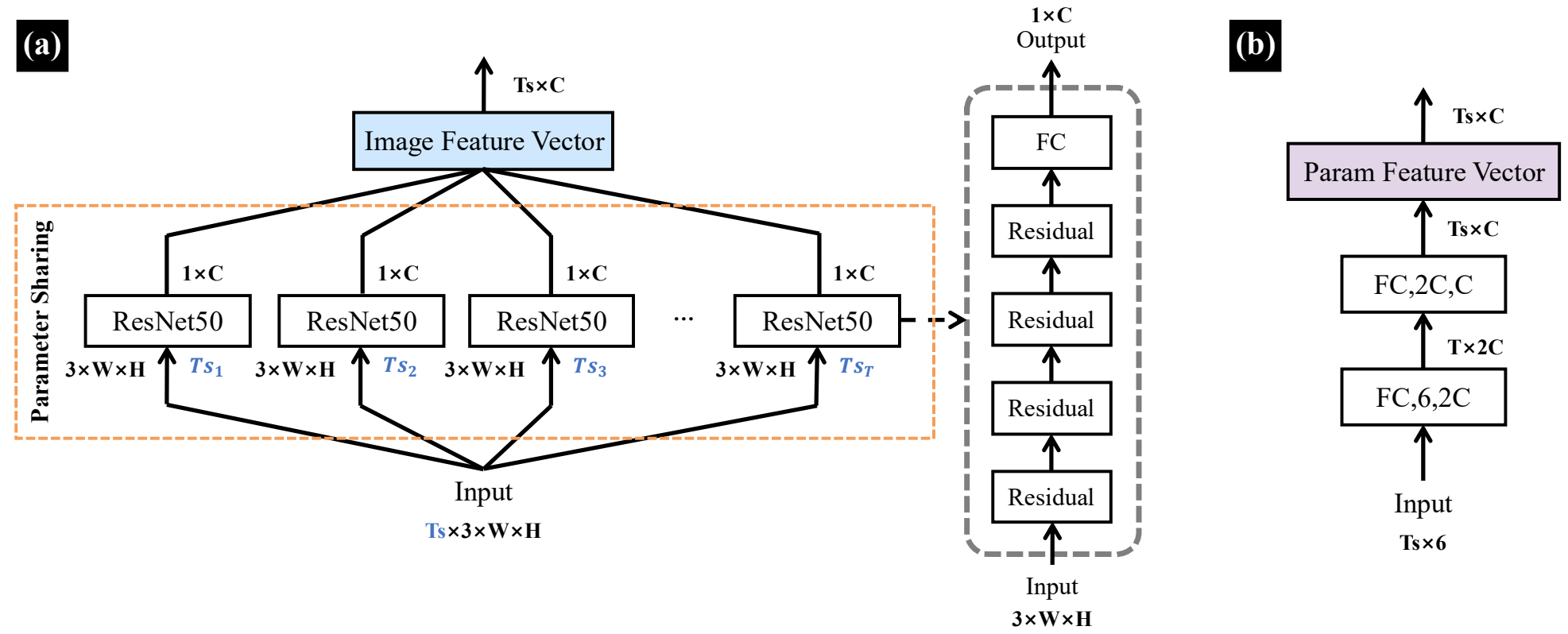

**Fig. 7** (a): the schematic diagram for the extraction of spatial features using ResNet50; (b): structural diagram for the extraction of welding parameter features.

To address this spatiotemporal decoupling, the ConvLSTM architecture was introduced (Shi et al., 2015). Its key innovation is the integration of convolutional operations within the LSTM gates, enabling it to directly process feature maps with spatial dimensions. As illustrated in **Fig. 8**, this allows the network to capture spatial and temporal information simultaneously. While this represents an improvement, the ConvLSTM's recurrent nature still constrains its ability to efficiently capture long-range dependencies, a critical requirement for modeling phenomena like heat accumulation over the entire welding process. Furthermore, ConvLSTM processes information through a fixed, local, and sequential update, lacking a mechanism to dynamically prioritize more distant but critical historical states.

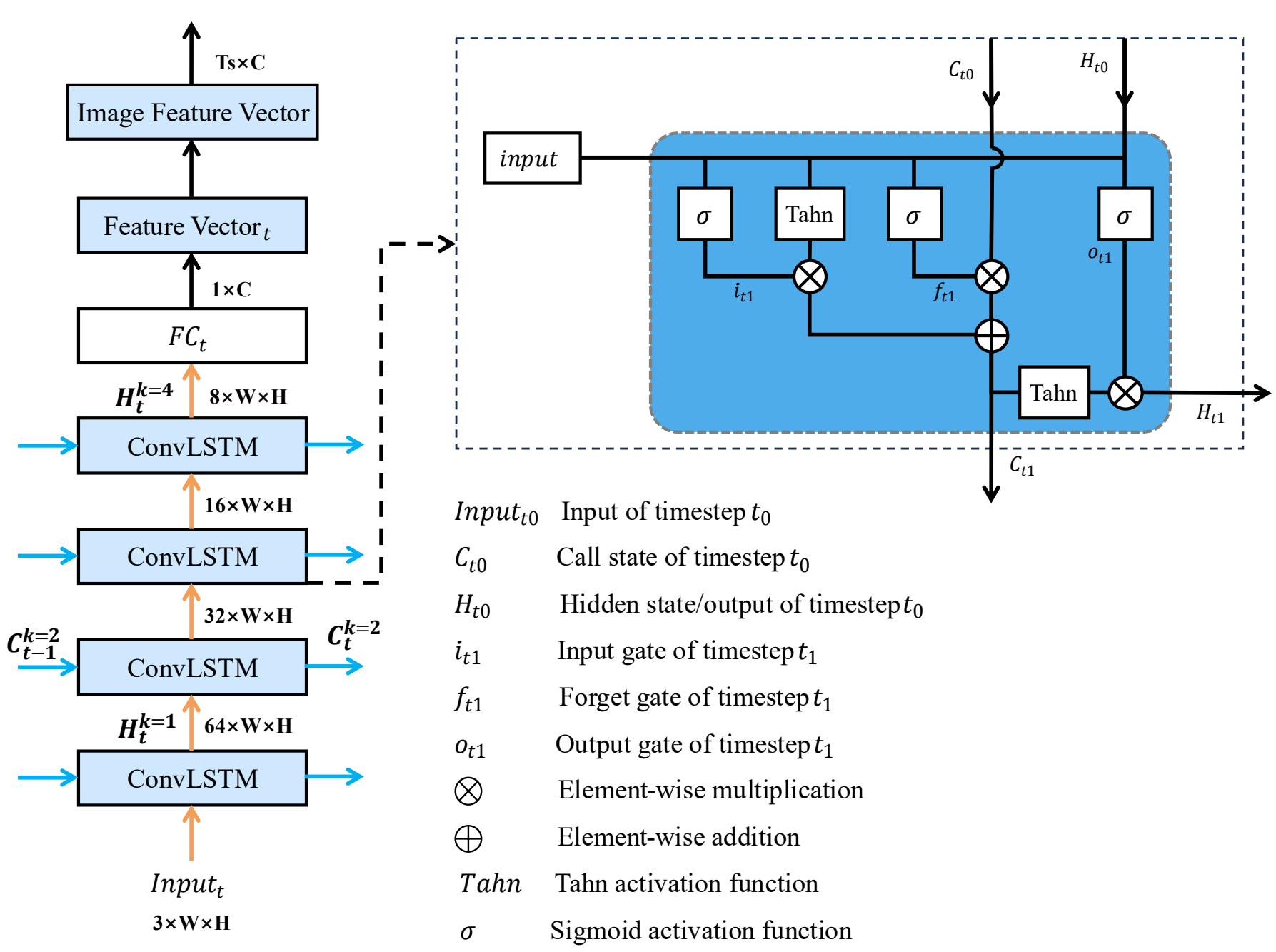

**Fig. 8** The structure of the predictive network with stacked ConvLSTMs. Right: The structure of the ConvLSTM.

To more effectively capture the long-range spatiotemporal dependencies and subtle motion changes inherent in the welding process, we employ a network based on the MAU. The selection of MAU is motivated by the limitations of conventional recurrent architectures like CNN-LSTM and ConvLSTM in modeling complex, non-sequential dynamics. While effective, they process information in a strictly sequential manner, which can be suboptimal for capturing the intricate fluid dynamics of the weld pool. The MAU is specifically designed to address this by incorporating two key mechanisms. As illustrated in **Fig. 9** , an MAU module takes two primary inputs at time step $t$ and layer $k$: the historical temporal states $T_{t-\tau:t-1}^{k}$ from the previous τ time steps, and the current spatial state $S_{t-1}^{k}$ from the layer below. Its operation unfolds through two sophisticated internal stages.

The process begins within its attention module, which addresses the challenge of efficiently utilizing historical information. It dynamically aggregates past temporal states by assigning importance weights, allowing the model to focus on the most relevant moments in the weld pool's history. First, a spatially-guided attention score $\alpha_j$ is calculated for each past temporal state $T_{t-j}^{k}$ based on its correlation with the current spatial state $S_t^{k-1}$. These scores are used to compute the long-term temporal information, $T_{att}$, as a weighted sum:

$$T_{att} = \sum_{j=1}^{\tau} \alpha_j \cdot T_{t-j}^{k} \tag{1}$$

This long-term information is then fused with the short-term temporal state ($T_{t-1}^{k}$) using a fusion gate $U_f$ to produce the augmented motion information, $T_{AMI}$:

$$U_f = \sigma\left(W_f * T_{t-1}^{k}\right); T_{AMI} = U_f \odot T_{t-1}^{k} + \left(1 - U_f\right) \odot T_{att} \tag{2}$$

where $\sigma$ denotes the sigmoid function, $*$ is the convolutional operator, and $\odot$ is the Hadamard (element-wise) product.

Once this augmented motion information ($T_{AMI}$) is generated, the module proceeds to its fusion stage. This second stage performs the core spatiotemporal interaction, using both $T_{AMI}$ and the current spatial state ($S_t^{k-1}$) to mutually update each other. This process is controlled by a temporal update gate ($U_t$) and a spatial update gate ($U_s$), ensuring a balanced fusion:

$$\begin{cases} U_t = \sigma(W_{tu} * T_{AMI}) \\ U_s = \sigma\left(W_{su} * S_{t-1}^{k}\right) \end{cases} \tag{3}$$

Using these gates, the new spatial state $S_t^k$ and temporal state $T_t^k$ are generated. This coupled update mechanism makes the MAU particularly suitable for capturing the intertwined dynamics of the weld pool:

$$S_t^k = f_{updates_S}\left(S_t^{k-1}, T_{AMI}, U_s\right); T_t^k = f_{update_T}\left(S_t^{k-1}, T_{AMI}, U_t\right) \tag{4}$$

Given the aforementioned details, the complete spatiotemporal feature extraction process can be concisely defined as:

$$S_t^0 = Enc(input_t) \tag{5}$$

$$Feature\ Vector_{\ t} = S_t^N \tag{6}$$

Herein, *Enc* denotes the initial encoder layer, $N$ signifies the number of stacked MAU layers (four in this research), and $input_t$ is the weld pool image at time *t*.

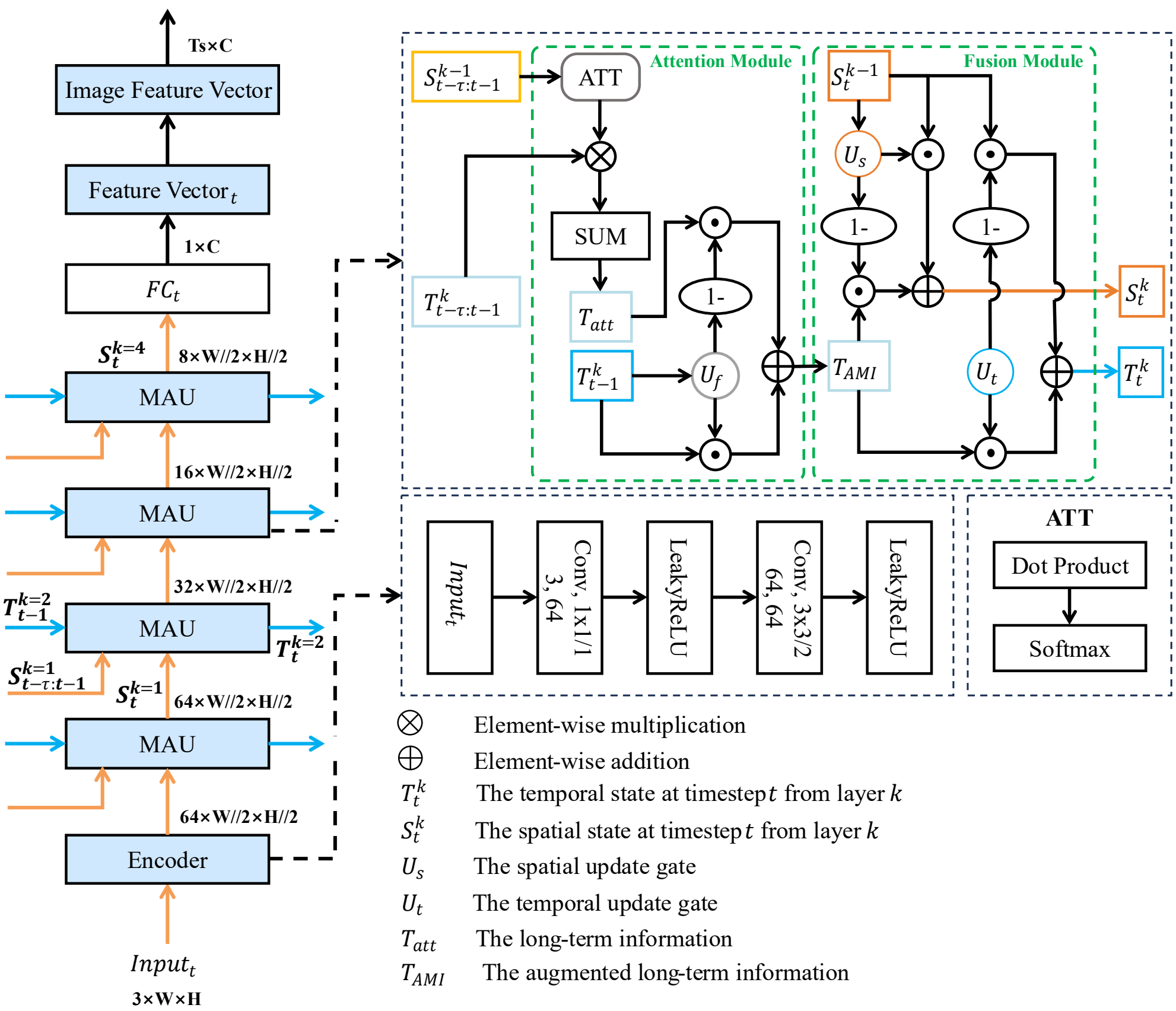

**Fig. 9** The structure of the predictive network with stacked MAUs. Right: The structure of the MAU.

### 3.2 Penetration state and depth prediction network (PSDPN)

Having extracted significant spatiotemporal features, the next challenge is to fuse them effectively and make predictions. For this, we chose a sophisticated fusion architecture based on the MFN (Zadeh et al., 2018) and leveraged Mamba(Rahman et al., 2024) as its core sequential processing engine. The MFN was selected to explicitly model the dynamic, cross-modal interactions between the visual data and the process parameters, a task for which simple fusion methods like concatenation are ill-suited. MFN is designed to learn the synergistic and temporal dependencies between modalities. It employs a dedicated gated memory component that continuously stores and updates a representation of the relationship between the two information streams. This enables the model to learn the direct physical cause-and-effect relationship between a change in a process parameter and the resulting visual dynamics of the weld pool, leading to a more physically grounded and accurate prediction.

Two baseline fusion strategies were first considered, as illustrated in **Fig. 10** and **Fig. 11**. The first, Type I (Early Fusion), involves concatenating the image and parameter feature sequences at the outset and processing the combined sequence with a single Bi-Mamba encoder. This approach is straightforward but forces the network to learn inter-modal relationships from a raw concatenation. The second, Type II (Late Fusion), processes each modality with a separate encoder before fusing them via a simple element-wise multiplication. While this allows for modality-specific feature learning, the final fusion step is shallow and may fail to capture complex, non-linear interactions. Both strategies, while functional, lack a dedicated mechanism for modeling the continuous, fine-grained cross-modal influences over time.

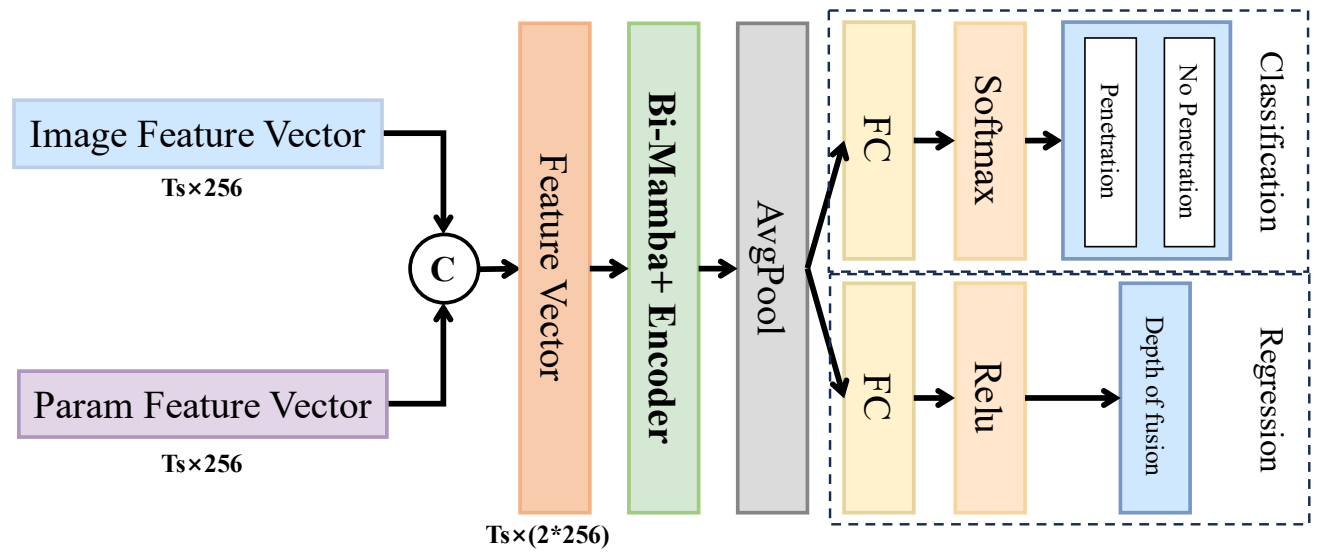

**Fig. 10** Spatiotemporal feature fusion network: Type I.

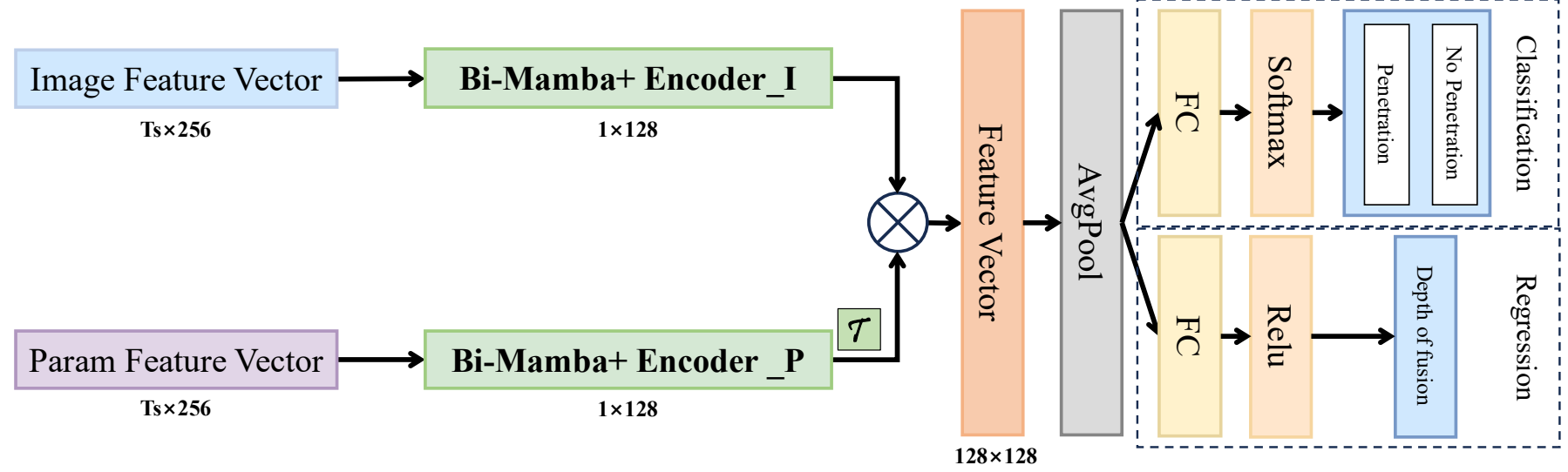

**Fig. 11** Spatiotemporal feature fusion network: Type II.

To overcome the limitations of these shallow fusion methods, we adopted an advanced architecture based on the MFN, which we adapted by integrating Mamba blocks for enhanced sequential processing (MFN-Mamba). The MFN-Mamba primarily consists of three components: 1) Mamba+ Block, which comprises multiple Mamba+ networks; 2) Delta-memory attention network, a specialized attention mechanism designed for detecting lateral and temporal interactions among different memory dimensions within the Mamba+ Block; 3) Multi-view gated memory, a unified memory for storing variations in images and welding parameters over time. Adjustments were made to adapt the MFN-Mamba network according to the characteristics of our model. The modified MFN-Mamba network structure is illustrated in **Fig. 12**.

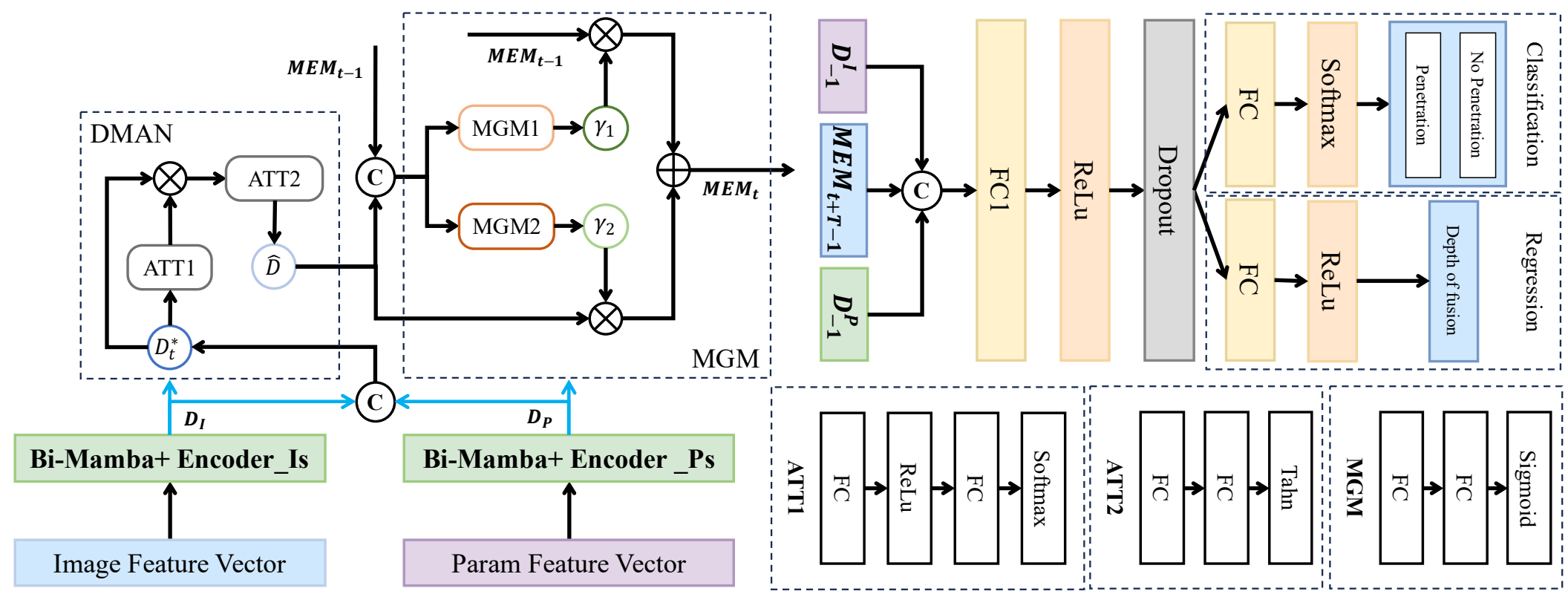

**Fig. 12** Spatiotemporal feature fusion network based on MFN-Mamba. The Mamba+ Block for both images and welding parameters consists of four stacked layers.

The primary function of the DMAN network is to integrate spatiotemporal features from two Mamba+ Blocks and facilitate their interaction at the time step $t$. To address this, a coefficient allocation technique is introduced, assigning large coefficients to the interaction features formed by the two spatiotemporal features, while smaller coefficients are allocated to other features. Additionally, DMAN can also recognize interactions occurring at different times, as its inputs comprise the outputs $D$ from the Mamba+ Block that carry spatiotemporal feature information across various time points.

MGM serves as a unified memory storage module for storing the history of interactions in conjunction with the Mamba+ Block. It is controlled using set of two gates. $\gamma_1$ (retain gate) determines how much of the current state of MGM should be remembered, while $\gamma_2$ (update gate) determines how much of MGM should be updated based on the update proposal $\hat{c}$ at each timestep $t$.

The output of MFN-Mamba consists of three concatenated vectors: the final state $MGM_{t+T-1}$, and the hidden dimension $D_{-1}^{I}$ and $D_{-1}^{P}$ from Mamba+ Block. Subsequent processing follows a consistent approach with previously proposed network methods.

### 3.3 Weld cross-section morphology reconstruction network (WCMRN)

The weld morphology cross-sectional reconstruction network employs the spatiotemporal features extracted in Section 3.2 to transform them into images representing the cross-sectional morphology of the weld bead. **Fig. 13** and **Fig. 14** respectively depict the flow diagram and structure of this network. The image generated by the decoder is denoted as $D_c$, while its corresponding actual value is represented as Truth. The penetration depth in the image generated is labeled as $Depth_c$, whereas the true penetration depth is referred to as $Depth_T$. It is important to clarify that our proposed framework employs a two-stage training strategy. In the first stage, the spatiotemporal feature extraction network (Section 3.1) and the penetration state and depth prediction network (PSDPN, Section 3.2) are trained jointly. This stage aims to optimize the feature extractor to effectively capture the spatiotemporal dynamics relevant to penetration state and depth. In the second stage, the weights of the pre-trained feature extraction network are frozen. This frozen encoder is then used to provide feature vectors to the weld cross-section morphology reconstruction network (WCMRN), which is trained independently. This approach leverages the knowledge learned from the classification and regression tasks to provide a robust feature foundation for the more challenging image generation task, improving training stability and final performance.

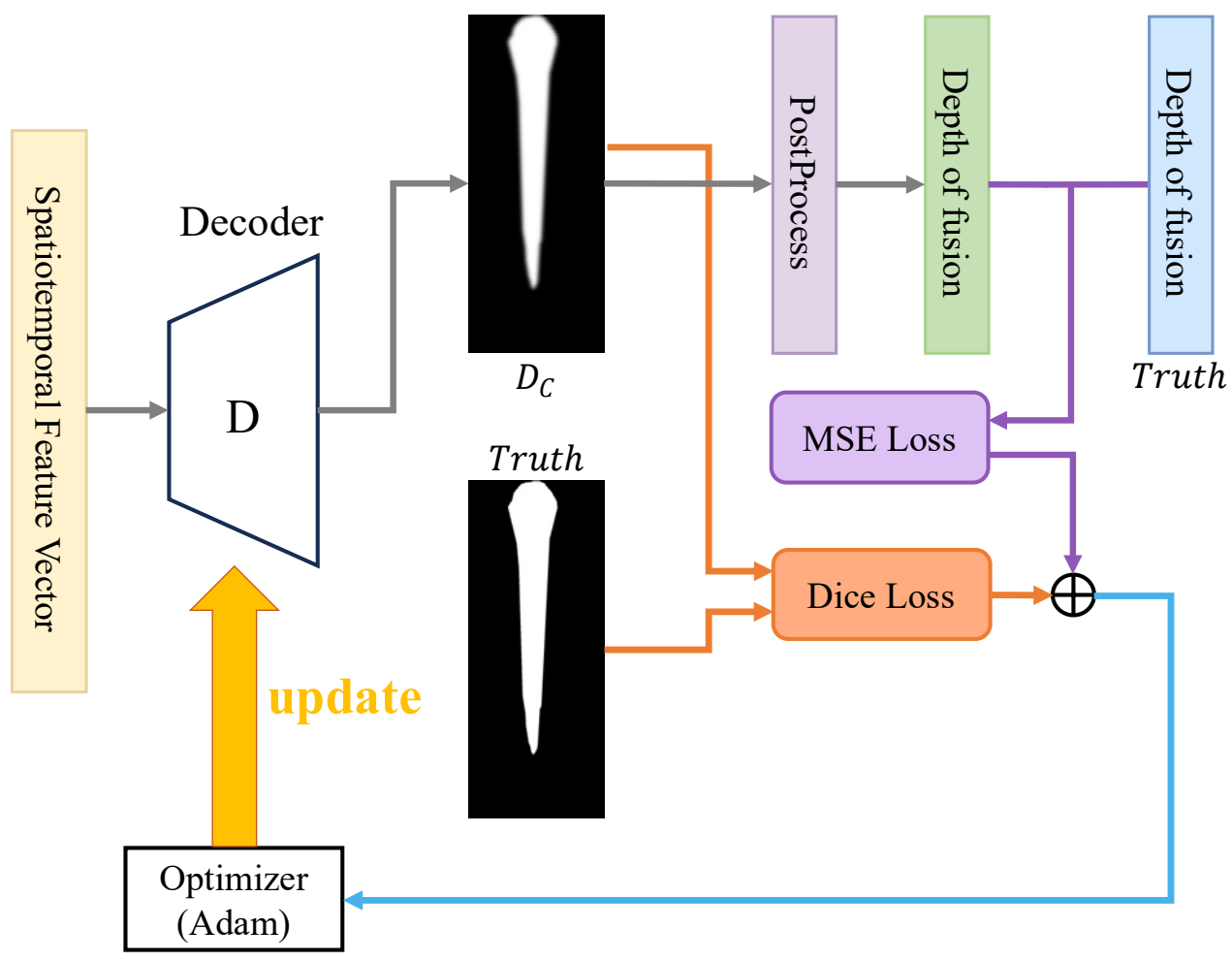

**Fig. 13** The training process diagram of the cross-sectional reconstruction network for weld beads. Process condition: $P_d = 9.0kW$; $P_r = 1.0kW$; $V = 1.5m/min$; $D_f = 0.0mm$; $T = 12.0mm, WP = 2$.

The overall pipeline, including the loss function and optimizer, is shown in **Fig. 13**. The weld cross-sectional reconstruction network primarily consists of a decoder network denoted as $D$ in the diagram. Mathematically, this network can be represented as:

$$D = D(c; \theta^D) \tag{7}$$

where, $\theta^D$ denotes the trained variables within $D$, encompassing weights and biases in the convolutional procedure. $\theta^D$ is trained to transform sequential features into the cross-sectional morphology of a weld bead. Our objective is to enhance the precision of the resulting cross-sectional morphology along the vertical axis, referred to as penetration depth, thus employing a composite loss function consisting of two parts: the first part, $J_1$ denotes the Dice loss between ground truths and neural network-generated cross-sectional morphology of the weld bead. The second part, $J_2$ represents mean square error between actual penetration depth and that inferred from neural network-generated cross-sectional morphology of the weld bead. The mathematical representation is derived as follow:

$$\begin{aligned} J^\theta(\theta^D) &= J_1 + \alpha \cdot J_2 \\ &= Dice\ Loss + \frac{\alpha}{n}\sum_{k}^{n}(Depth_T^k - Depth_C^k)^2 \end{aligned} \tag{8}$$

where, $\alpha$ is defined as a weight function that regulates the degree of significance for $J_2$ loss function, and it incrementally increases throughout the entire training process. Subsequently, the entire network's weights are updated by the Adam(Kingma and Ba, 2017) optimizer based on $J^\theta(\theta^D)$.

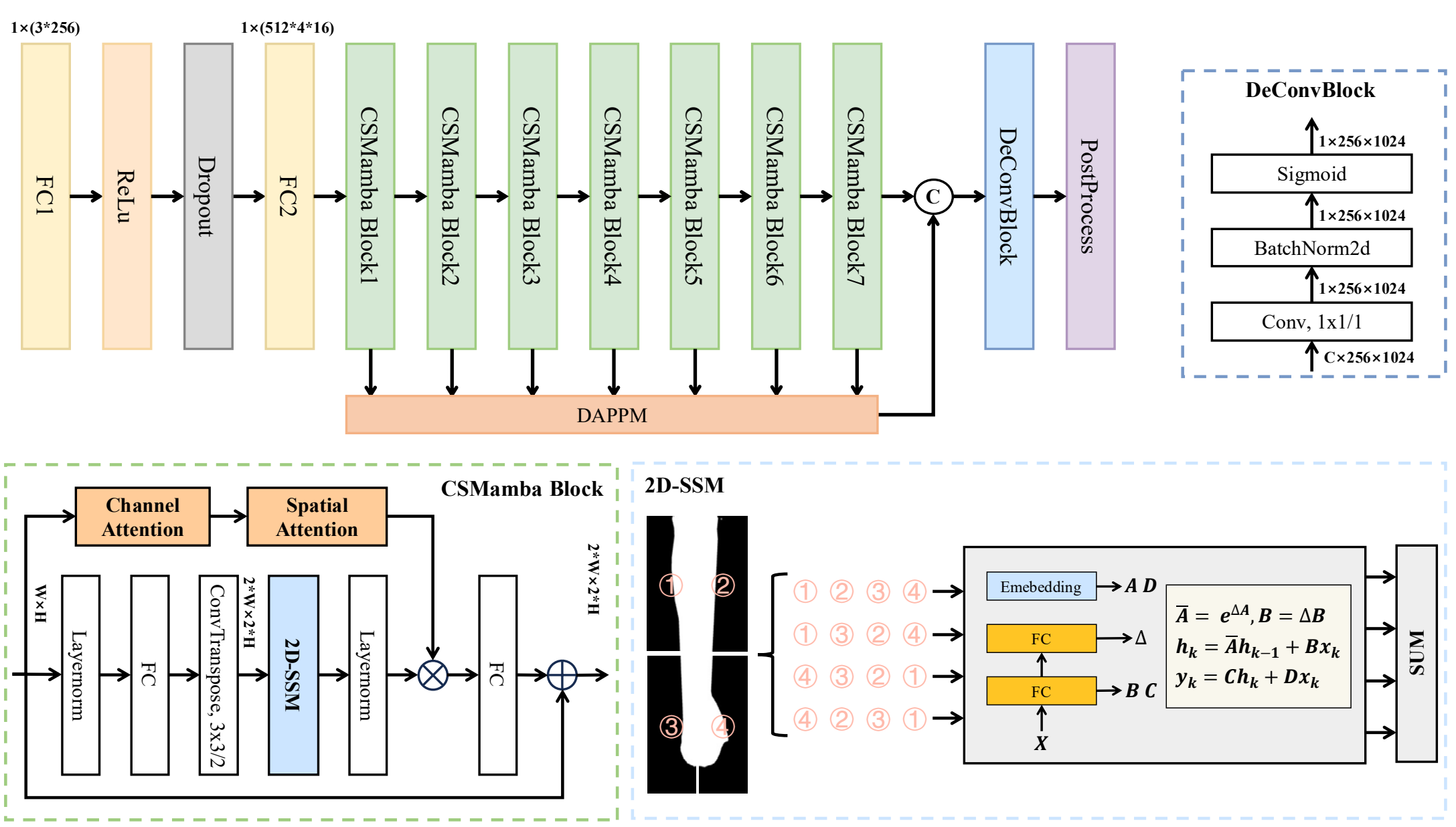

**Fig. 14** Detailed architecture of weld cross-section morphology reconstruction network (WCMRN).

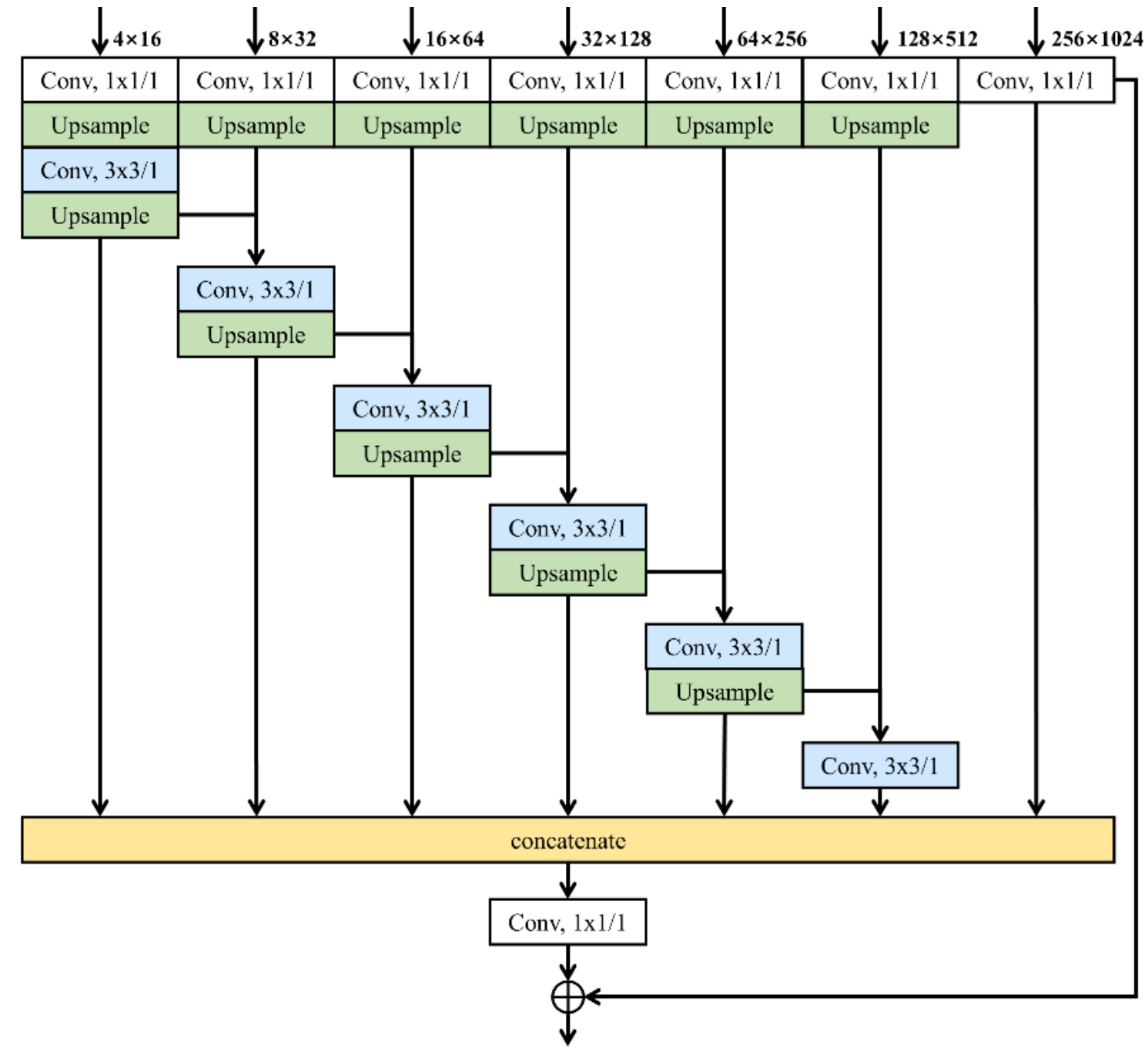

**Fig. 15** Detailed architecture of deep aggregation pyramid pooling module (DAPPM).

**Fig. 14** illustrates the detailed structure of the weld cross-section reconstruction network. The initial three layers are derived from the spatiotemporal sequence fusion network. The FC2 layer reshapes the $1 \times 256 * 3$ feature map into a $1 \times 512 * 4 * 16$ format, which is then expanded to a $4 \times 256 \times 1024$ shape via seven CSMamba(Liu et al., 2024) blocks. To more effectively extract contextual information from low-resolution feature maps, we incorporate the deep aggregation pyramid pooling module (DAPPM), inspired by Hong's work (Hong et al., 2021). Modifications have been made to adapt this module to our model's structural characteristics. Within each ConvTranspose block, the generated feature maps are fed into DAPPM for further processing.

Initially, low-resolution images entering DAPPM undergo channel adjustment using a $1 \times 1$ convolution module and subsequent up-sampling of the feature map followed by fusion of multiscale contextual information in a hierarchical-residual manner using additional $3 \times 3$ convolution modules. All resulting features are then concatenated and subjected to channel adjustment via a $1 \times 1$ convolution operation. Finally, the output feature map is concatenated with the one generated by the last CSMamba block and passed on to DeConvBlock for conversion into a grayscale image. The PostProcessBlock converts the grayscale image obtained above into a binary image and calculates the number of vertical pixels. This pixel count along with real-time actual penetration depth is used to compute $J_2$ portion of the loss function after converting pixel values to penetration depth values based on calibration data during training stage only, it remains disengaged during validation and testing stages.

### 3.4 Valuation index

The binary classification model used in this study necessitates the utilization of four metrics for penetration prediction: true positives (TP), false positives (FP), false negatives (FN), and true negatives (TN). Furthermore, the following four evaluation indicators are utilized to quantitatively assess the presented model in this paper, namely:

$$Accuracy\ (Acc) = \frac{TP + TN}{TP + TN + FP + FN} \times 100\% \tag{9}$$

$$Precision\ (P) = \frac{TP}{TP + FP} \times 100\% \tag{10}$$

$$Recall\ (R) = \frac{TP}{TP + FN} \times 100\% \tag{11}$$

$$F1_{score}(F1) = 2 \times \frac{P \times R}{P + R} \tag{12}$$

For the penetration depth prediction model, the RMSE is utilized as a metric to evaluate the prediction accuracy of the model. It computationally determines the average of the squares of the differences between the predicted and actual values. Its mathematical expression is:

$$RMSE = \sqrt{\frac{1}{n}\sum_{i=1}^{n}(y_i - \hat{y}_i)^2} \tag{13}$$

The introduction of the R-squared metric was utilized to characterize the model's performance. This indicator represents the model's explanatory power regarding the dependent variable, with a value range from 0 to 1. A value closer to 1 indicates a better fit to the data. Its mathematical expression is:

$$R^2 = 1 - \frac{SS_{res}}{SS_{tot}} = 1 - \frac{\sum_{i=1}^{n}(y_i - \hat{y}_i)^2}{\sum_{i=1}^{n}(y_i - \bar{y})^2} \tag{14}$$

Here, $y_i$ represents the actual values, $\hat{y}_i$ denotes the predicted values from the model, and $\bar{y}$ is the mean of the actual values. In addition, the MAE was introduced, which calculates the average of the absolute values of the prediction errors for each sample. MAE places more emphasis on the magnitude of the disparity between the model's predictions and the true values. It disregards the direction of the prediction errors and focuses on their absolute magnitude. The mathematical expression for MAE is:

$$MAE = \frac{1}{n}\sum_{i=1}^{n}|y_i - \hat{y}_i| \tag{15}$$

The weld cross-sectional reconstruction network employs intersection over union (IoU) and dice coefficient to assess the performance of the model, their mathematical expressions are:

$$IoU = \frac{|A \cap B|}{|A \cup B|} \tag{16}$$

$$Dice = \frac{2|A \cap B|}{|A| + |B|} \tag{17}$$

where $A$ and $B$ represent the predicted region and the actual region, respectively. The notation $|A \cap B|$ denotes the part accurately predicted by the model, $|A \cup B|$ signifies the union of all predicted regions and actual regions, and $|A| + |B|$ indicates the sum of the areas of all predicted and actual regions.

For the binary classification task of penetration state, the Receiver Operating Characteristic (ROC) curve is plotted to visualize the model's performance across all classification thresholds. The Area Under the Curve (AUC) is calculated as a single scalar value representing the overall classification capability.

For the weld cross-section reconstruction task, the HD95 is employed to evaluate the boundary similarity. Unlike the standard Hausdorff Distance which is sensitive to outliers, HD95 is calculated by computing the distance from each point on one boundary to the closest point on the other, and then taking the 95th percentile of these distances. This provides a more stable measure of the discrepancy between the two contours.

### 3.5 Training information

The learning rate during training and the exponential decay rate for the first moment estimate in the Adam optimizer ($\beta_1$) is set at $1 \times e^{-3}$ and 0.9, respectively. To mitigate the adverse impact of a large initial learning rate on training, a warmup scheme is employed, gradually increasing the learning rate from 0 to $1 \times e^{-3}$ over the first three epochs. This approach helps prevent oscillations in the early stages of model training. The learning rate decay follows a polynomial decay scheme, with its variations over the training epochs depicted in **Fig. 16**. During training, the $TSL$ input to the model is set as 50, with a prediction time of 1 second. Consequently, the network utilizes 50 top surface weld pool images and welding parameters from time $(t-1)\ s$ to $t\ s$ to predict penetration status, depth of penetration, and cross-sectional morphology at time $t\ s$. The model is trained with a batch size of 150. To ensure robust generalization and prevent overfitting, we save the model checkpoint that achieves the best performance on the validation set. This saved model is then used for a final evaluation on the test set, and all reported results are based on its performance on this test set.

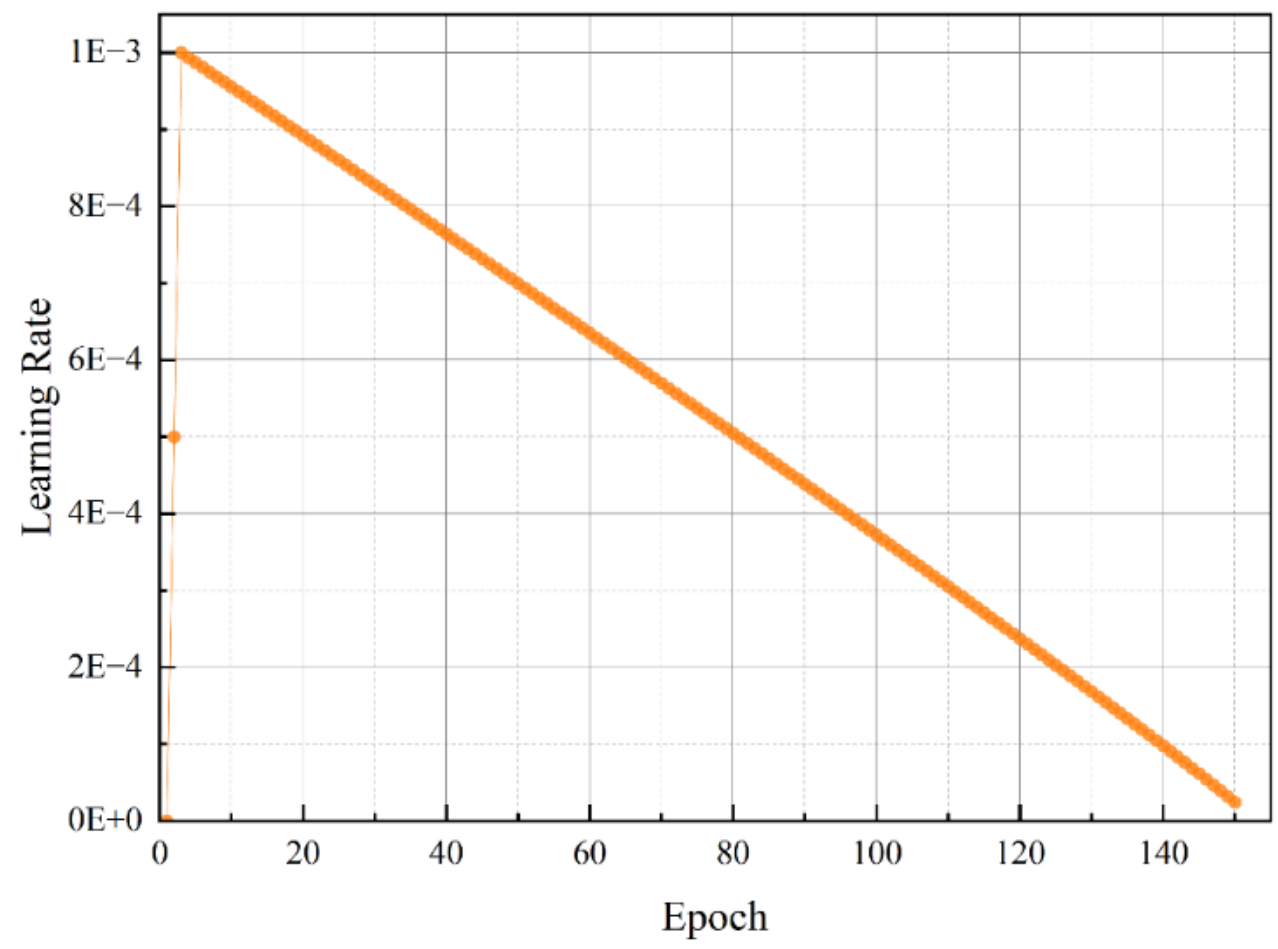

**Fig. 16** The variation curve of the learning rate during the training process.

# 4. Results and discussion

## 4.1 Ablation experiment

The training was conducted based on the network architectures proposed in Section 3, utilizing the methodologies outlined in Section 1. Initially, the study involved training and performing a comparative analysis of classification networks consisting of three spatiotemporal feature extraction networks and three spatiotemporal feature fusion networks. **Fig. 17** illustrates the process of enhancing accuracy while reducing error during the training. **Fig. 18** illustrates the ROC curves for models. **Tab. 5** presents the performance evaluation results of all nine network architectures on the validation set. To provide a broader comparison, we also evaluated several other deep learning models. These include TCN(Bai et al., 2018), TimeSformer(Bertasius et al., 2021) and Pyraformer(Liu et al., 2021), with their comparative results included in the same table.

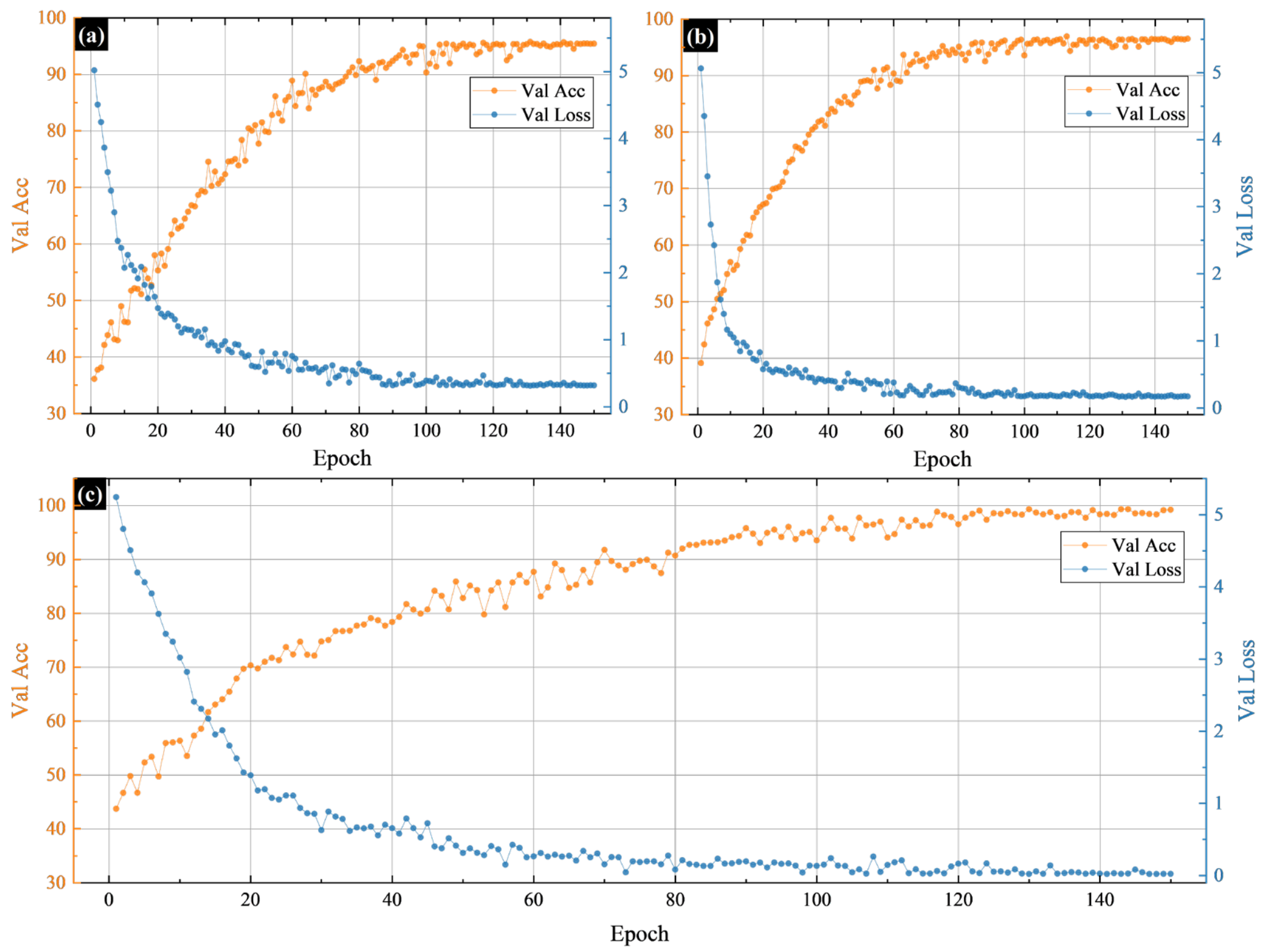

**Fig. 17** Training process of models: (a) CNN-LSTM + MFN-Mamba; (b) ConvLSTM + MFN-Mamba; (c) MAU + MFN-Mamba.

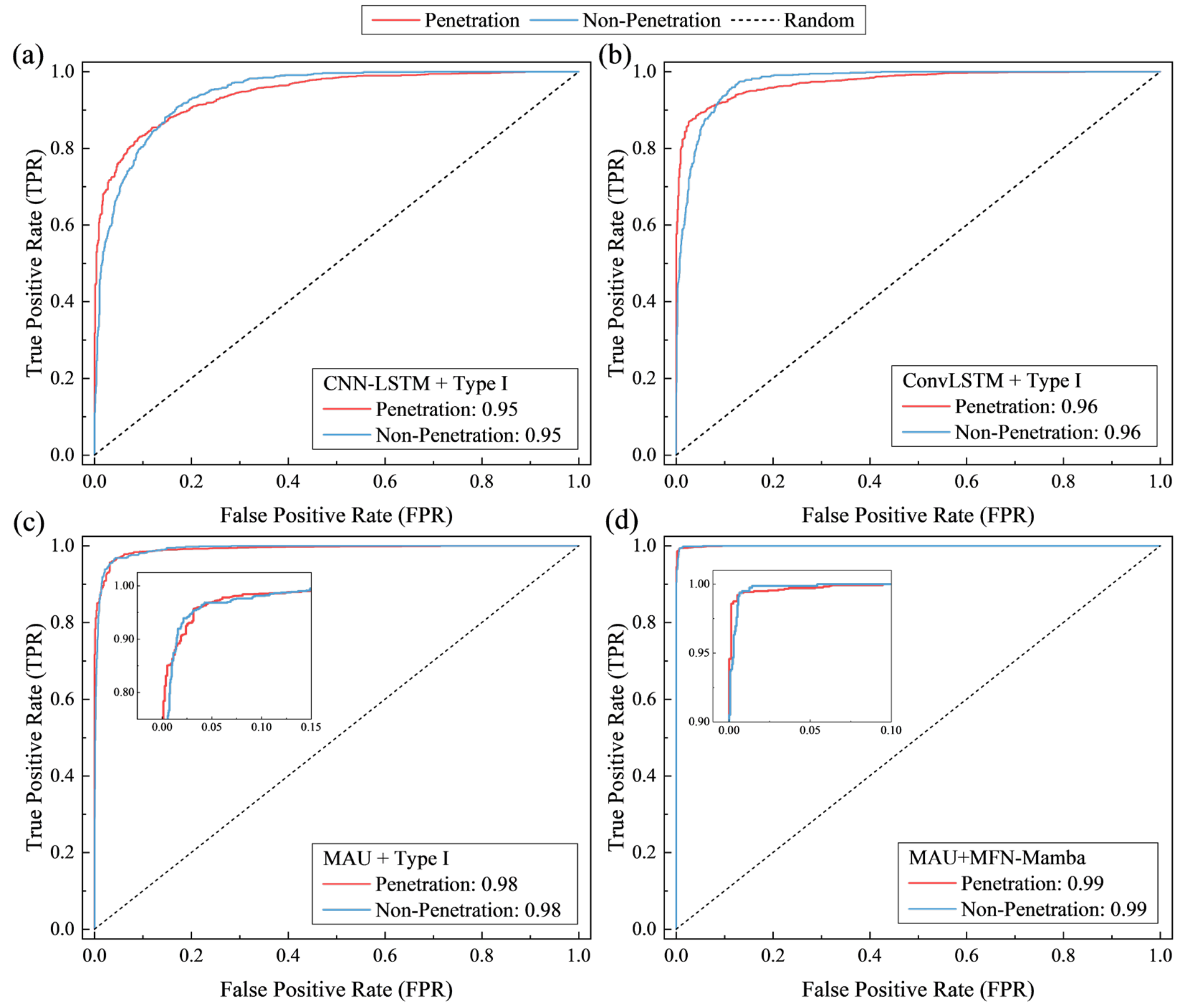

**Fig. 18** ROC Curves for the Models: (a) CNN-LSTM + Type I; (b) ConvLSTM + Type I; (c) MAU + Type I; (d) MAU+MFN-Mamba.

**Tab. 5** Performance of penetration prediction models in test set. Best results are displayed in bold and the runner-up results are underlined.

| No. | Model (feature extraction network + feature fusion network) | $Acc(\%)\uparrow$ | $P(\%)\uparrow$ | $R(\%)\uparrow$ | $F1(\%)\uparrow$ | Inference time(ms)↓ |
|---|---|---|---|---|---|---|
| 1 | CNN +TCN + MFN-Mamba | 98.15 | 98.20 | 98.02 | 98.11 | 95.80 |
| 2 | TimeSformer + MFN-Mamba | <u>98.67</u> | <u>98.54</u> | <u>98.76</u> | <u>98.65</u> | 140.15 |
| 3 | CNN + Pyraformer + MFN-Mamba | 97.90 | 98.01 | 97.71 | 97.86 | 129.43 |
| 4 | CNN-LSTM + Type I | 87.15 | 81.84 | 90.76 | 86.06 | **40.77** |
| 5 | CNN-LSTM + Type II | 90.93 | 88.75 | 93.16 | 90.90 | 41.89 |
| 6 | CNN-LSTM+MFN | 91.79 | 92.86 | 92.13 | 92.49 | <u>50.66</u> |
| 7 | CNN-LSTM+MFN-Mamba | 95.63 | 95.00 | 96.27 | 95.63 | 63.15 |
| 8 | ConvLSTM + Type I | 91.58 | 88.09 | 87.48 | 87.78 | 69.41 |
| 9 | ConvLSTM + Type II | 92.71 | 91.49 | 94.00 | 92.72 | 71.96 |
| 10 | ConvLSTM + MFN | 94.40 | 93.72 | 95.07 | 94.39 | 80.68 |
| 11 | ConvLSTM + MFN-Mamba | 96.00 | 94.67 | 95.62 | 95.14 | 89.02 |
| 12 | MAU + Type I | 96.17 | 96.55 | 96.70 | 96.62 | 66.75 |
| 13 | MAU + Type II | 97.34 | 98.26 | 97.31 | 97.78 | 75.44 |
| 14 | MAU+MFN | 98.18 | 97.91 | 98.32 | 98.11 | 90.14 |
| 15 | **MAU+MFN-Mamba** | **99.35** | **99.35** | **99.35** | **99.35** | 105.45 |

The results presented in **Tab. 5** lead to a clear conclusion: the model utilizing the MAU+MFN-Mamba combination (No. 15) achieves the highest overall performance. This superiority stems from a carefully designed architecture, as demonstrated by the ablation study. Firstly, a comparison of the feature extraction networks (e.g., No. 7, No. 11, and No. 15, all using the same MFN-Mamba fusion) unequivocally demonstrates that MAU exhibits superior proficiency in extracting spatiotemporal features from images when compared to CNN-LSTM and ConvLSTM. Secondly, an analysis of the fusion networks (e.g., No. 12-15, all using the same MAU extractor) proves that MFN-Mamba is more effective at integrating image features with welding process parameters than the simpler Type I and Type II methods.

Having established our proposed combination as the optimal internal architecture, we further benchmarked it against several other powerful, general-purpose deep learning models (No. 1-3). It is noteworthy that our specialized model also outperforms these advanced architectures. This suggests that for this specific industrial application, an architecture like MAU, which is explicitly designed to be motion-aware, provides a tangible advantage over more general-purpose models for sequence analysis.

Consequently, the MAU+MFN-Mamba model represents the optimal choice for penetration state classification. Simultaneously, we evaluated the inference speed of the models. All training and inference tests were conducted on the same workstation, equipped with an NVIDIA GeForce RTX 2080 Ti GPU, which has a single-precision (FP32) performance of 13.45 TFLOPs. The results indicate that the inference time of the model is less than 200ms, which demonstrates its capability to perform real-time inference on weld penetration during the welding process. This serves as a critical prerequisite for the subsequent closed-loop control of laser welding.

To further investigate the impact of welding parameters on model performance, ablation experiments were conducted. **Fig. 19** illustrates the variations in accuracy and error rates observed during several ablation experiments. **Tab. 6** presents the influence of each welding parameter on model accuracy. A "+" indicates that the parameter was included in training, while a "-" indicates its exclusion. The table reveals that when only top surface weld pool images are used, resulting in a model structure containing only MAU without MFN-Mamba, the model's accuracy is significantly reduced to 89.4%, accompanied by notable fluctuations during training process. However, incorporating various welding parameters leads to improvements in model accuracy, with $P_d$ having the most significant effect followed by $P_r$. In dot-ring laser welding, penetration depth primarily depends on the dot laser while adding a ring laser enhances spot laser absorption rate and consequently improves penetration capability. Welding position has minimal impact on model accuracy as its inclusion results in merely a marginal improvement of 0.05%.

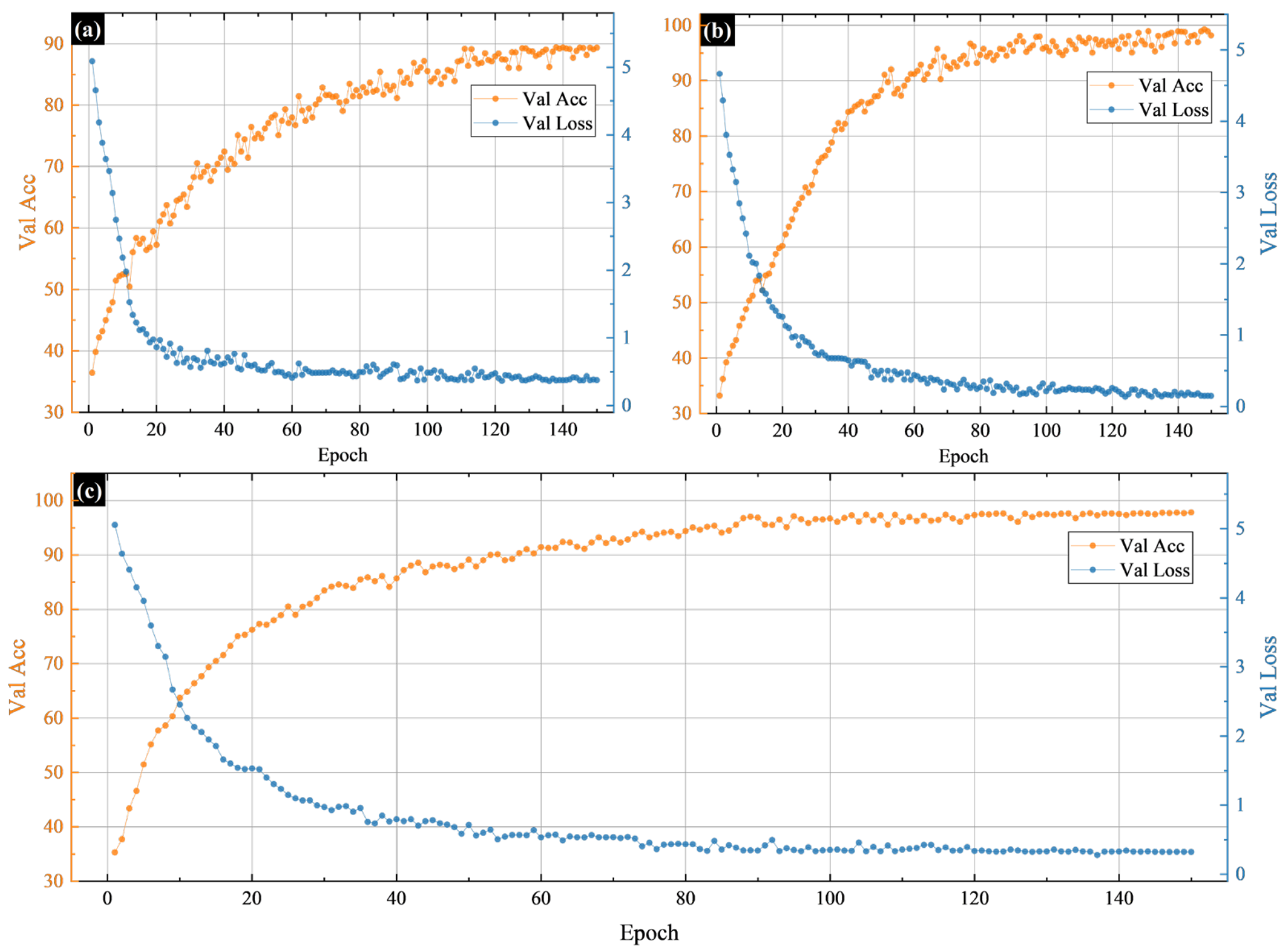

**Fig. 19** The training process of models: (a) Model No.1 (b) Model No.4(c) Model No.6

**Tab. 6** Ablation results of the penetration prediction network. Best results are displayed in bold and the runner-up results are underlined.

| No. | Welding parm | | | | | | $Acc(\%)\uparrow$ | $F1(\%)\uparrow$ |
|---|---|---|---|---|---|---|---|---|
| | $P_d$ | $P_r$ | $V$ | $D_f$ | $T$ | $WP$ | | |
| 1 | - | - | - | - | - | - | 89.46 | 89.17 |
| 2 | + | - | - | - | - | - | 93.28(+3.82) | 93.09(+3.92) |
| 3 | + | + | - | - | - | - | 96.32(+3.04) | 95.97(+2.88) |
| 4 | + | + | + | - | - | - | 97.71(+1.39) | 97.55(+1.58) |
| 5 | + | + | + | + | - | - | 98.69(+0.98) | 98.60(+1.05) |
| 6 | + | + | + | + | + | - | <u>99.30(+0.61)</u> | <u>99.28(+0.68)</u> |
| 7 | + | + | + | + | + | + | **99.35(+0.05)** | **99.30(+0.02)** |

Based on the results obtained from the classification network, we proceeded to train the depth of penetration prediction network. The spatiotemporal feature extraction network employed the three previously proposed networks for extracting spatiotemporal feature, while the feature fusion network utilized MFN-Mamba. All six welding process parameters were incorporated into the model training. **Tab. 7** illustrates how different spatiotemporal feature extraction networks, including the other advanced models mentioned previously, impact model accuracy. Similar to the performance of the classification model, our validation set showed that combining MAU with MFN-Mamba yielded superior results.

### 4.2 The accuracy and performance of PSDPN

It is easy to find from **Tab. 7** that the MAU + MFN-Mamba model which was proposed before has the best performance in predicting penetration depth. The penetration depth prediction results and correlative relation were shown in **Fig. 20**. It is not hard to see that the distribution of measured/predicted points was close to the perfect fit line. In order to further compare and analyze the models, multiple evaluation criteria were comprehensively considered using the Taylor diagram in **Fig. 21**. Compared to a single model evaluation metric, the Taylor diagram(Taylor, 2001) provides a more intuitive comparison of model performance. It effectively displays diverse model information and is widely used for model evaluation and testing. As evident from **Fig. 21**, the MAU + MFN-Mamba model shows the best performance in predicting penetration depth among all models.

The visualization of results for the three models on the test set is presented in **Fig. 22**. In this test data, the MAU + MFN-Mamba model achieves an average error of 0.41 mm. Notably, the figure indicates that the CNN-LSTM + MFN-Mamba model performs particularly poorly at the beginning of the welding process, where stability is at its lowest. Although the ConvLSTM + MFN-Mamba model can track the increase in penetration depth, its average error remains relatively high. The MAU + MFN-Mamba model successfully tracks the rapid changes in penetration depth. To our knowledge, this is particularly challenging, as only minimal information is available for predicting penetration depth amidst such dynamic fluctuations. In laser penetration welding with a thickness exceeding 8 mm, an error of 0.41 mm is considered acceptable.

**Tab. 7** Experimental results of the penetration depth prediction network. Best results are displayed in bold and the runner-up results are underlined.

| No. | Model (feature extraction network + feature fusion network) | $RMSE(mm)\downarrow$ | $R^2\uparrow$ | $MAE(mm)\downarrow$ | Inference time(ms)↓ |
|---|---|---|---|---|---|
| 1 | CNN + TCN + MFN-Mamba | 2.8341 | 0.9721 | 1.7224 | 96.10 |
| 2 | TimeSformer + MFN-Mamba | <u>2.1895</u> | <u>0.9790</u> | <u>1.3578</u> | 138.55 |
| 3 | CNN + Pyraformer + MFN-Mamba | 2.6112 | 0.9733 | 1.7056 | 126.20 |
| 4 | CNN-LSTM + MFN-Mamba | 4.8541 | 0.9135 | 2.5053 | **64.45** |
| 5 | ConvLSTM + MFN-Mamba | 3.6371(-1.2170) | 0.9666(+0.531) | 1.8818(-0.6235) | <u>89.15(+24.70)</u> |
| 6 | **MAU + MFN-Mamba** | **1.7914(-1.8457)** | **0.9814(+0.148)** | **1.0509(-0.8309)** | 107.41(+18.26) |

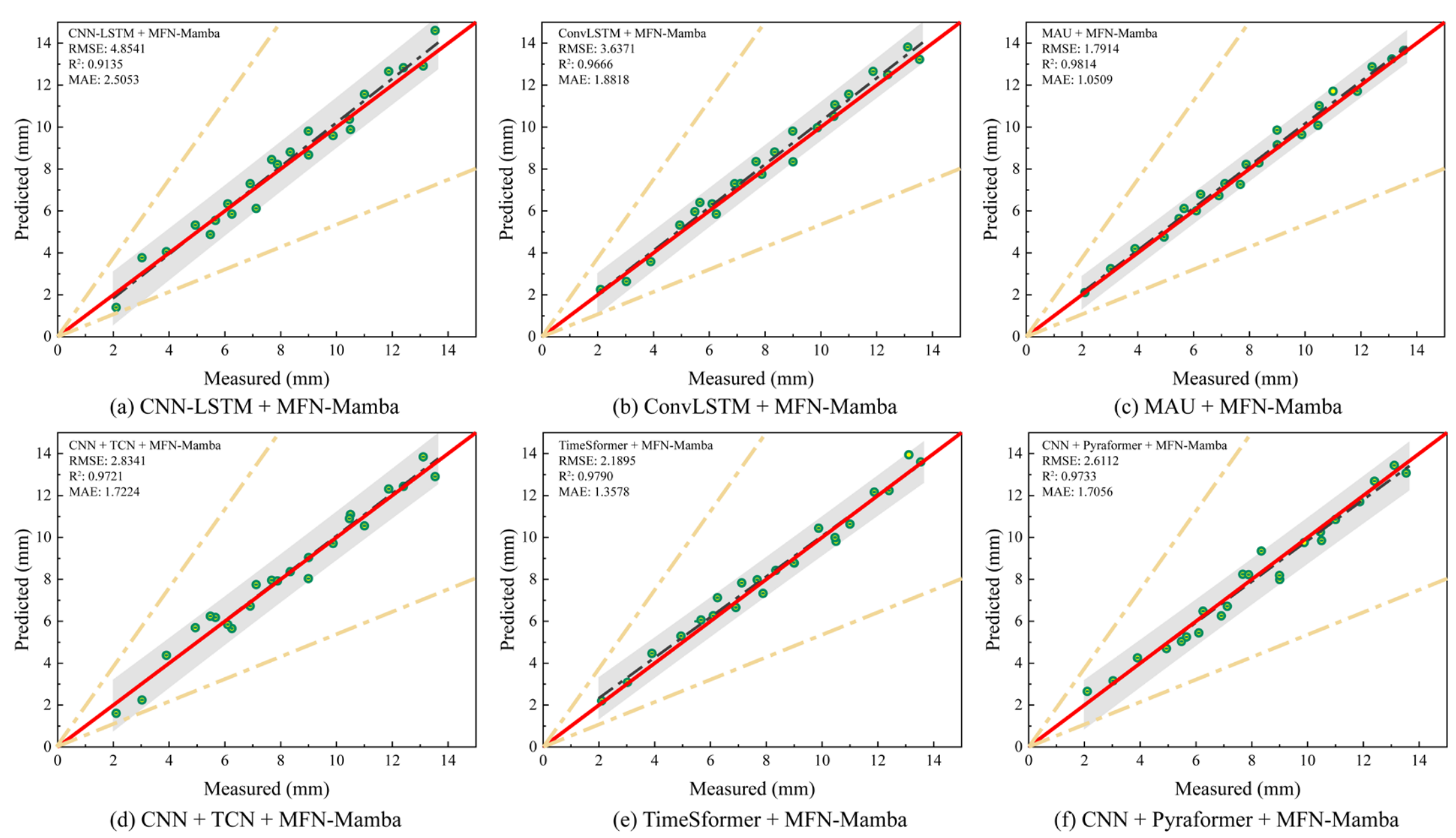

**Fig. 20** Correlation analyses of the measured and predicted values

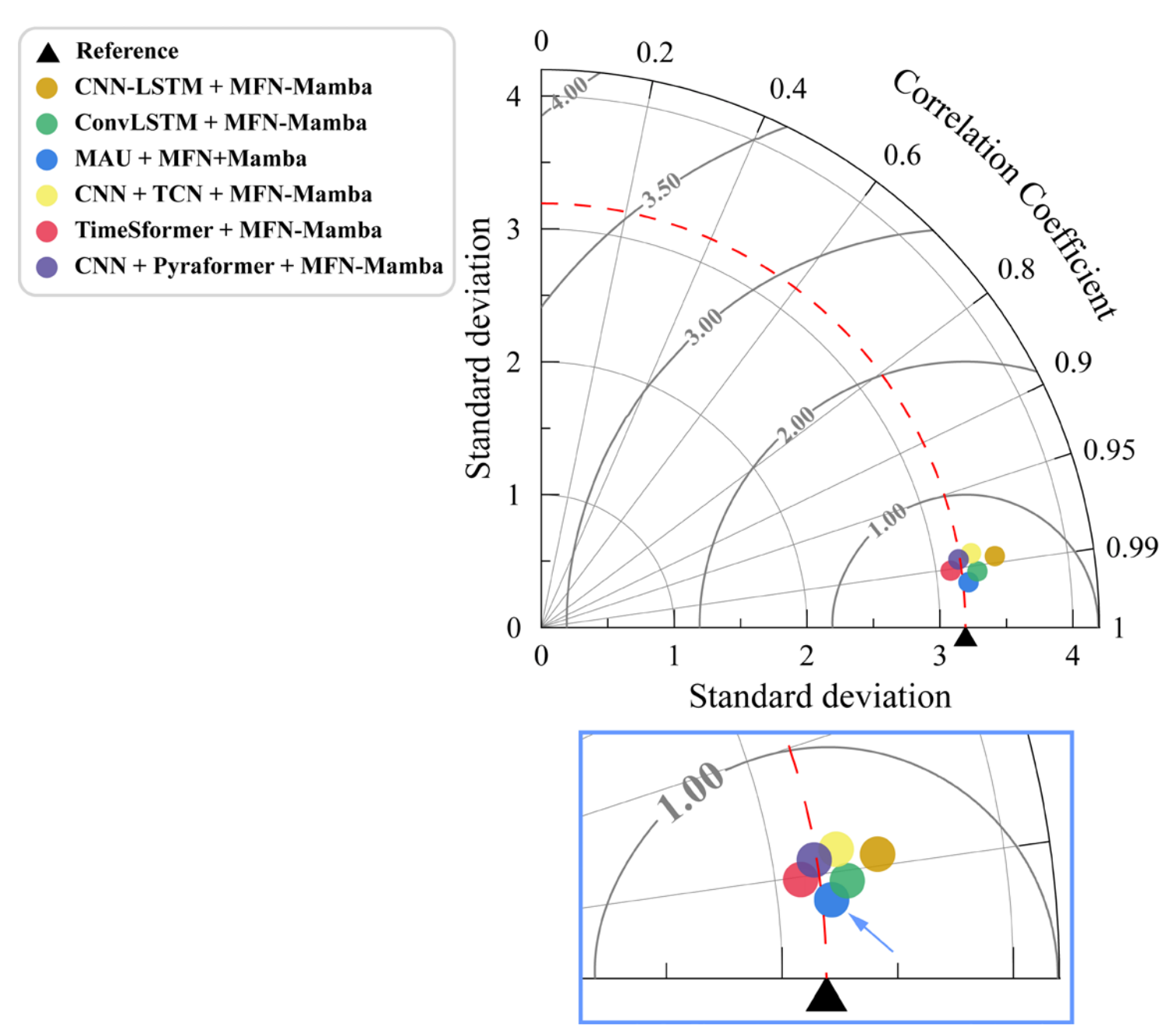

**Fig. 21** Performance comparison using Taylor diagram

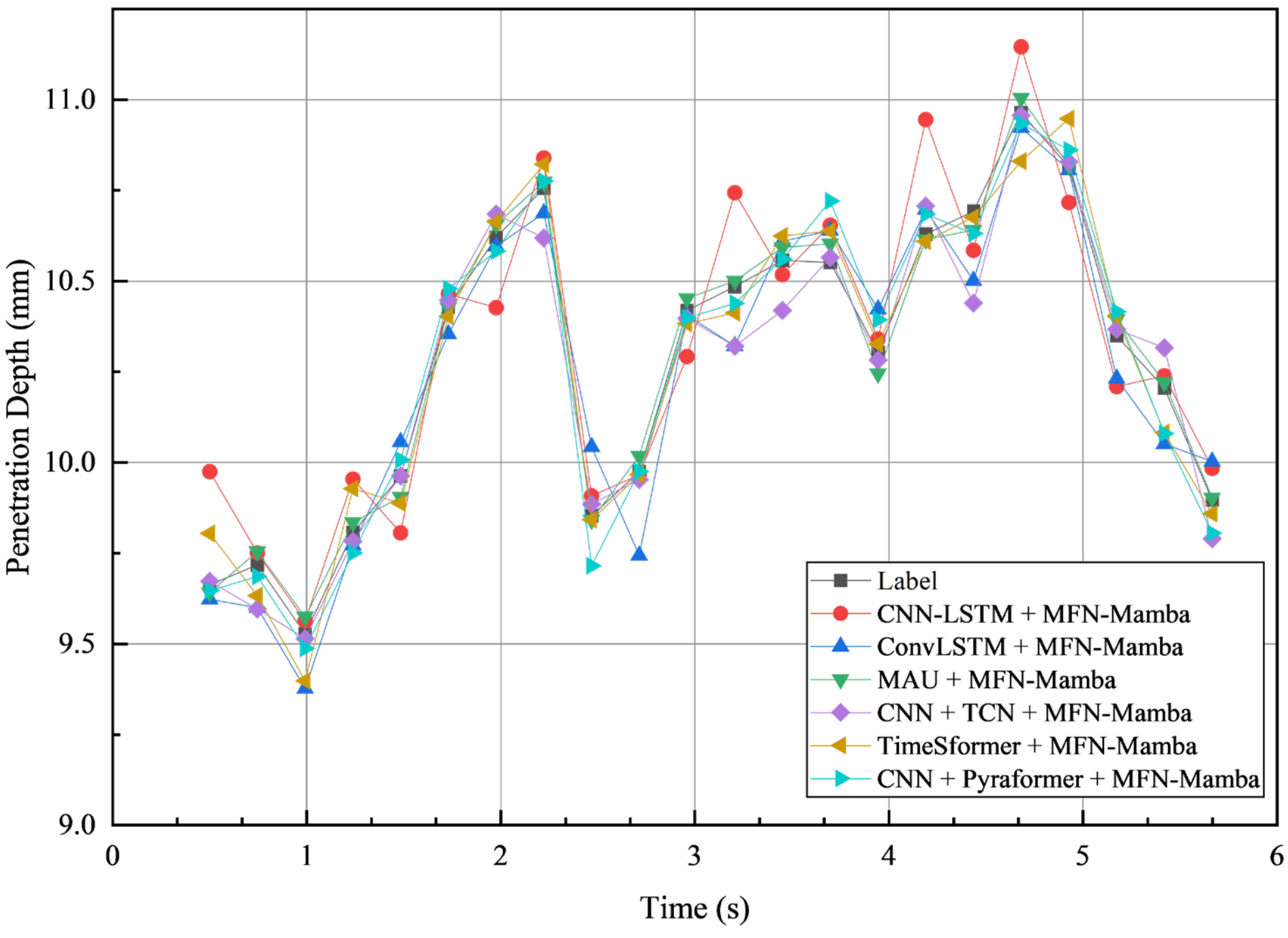

**Fig. 22** The penetration depth prediction network performance on test set.

### 4.3 The accuracy and performance of WCMRN

**Tab. 8** illustrates the impact of different spatiotemporal feature extraction networks on model accuracy. In order to evaluate the performance of the model, the inference results were visually presented. The prediction outcomes on the validation set are displayed in **Fig. 23**. The first-row exhibits images of the weld cross-section at various time points during the welding process. The second row shows the ground truth, followed by the inference results of the "CNN-LSTM + MFN-Mamba" model in the third row, the "ConvLSTM + MFN-Mamba" results in the fourth row, and the "MAU + MFN-Mamba" results in the fifth row. The final row demonstrates the performance of "MAU + MFN-Mamba" model with a novel loss function. The visualization reveals that models utilizing CNN-LSTM and ConvLSTM tend to generate symmetrical weld cross-sections but fail to accurately restore weld bead width. On contrary, employing MAU yields superior restoration of cross-section compared to previous models, particularly for reconstructing top surface bead width. However, penetration depth restoration remains inadequate as it often leads to either insufficient or excessive reconstruction. By applying our new loss function, we observe an improvement in penetration depth reconstruction ability which reduces discrepancies between reconstructed images and ground truth.

In order to enhance the performance verification of our proposed model, we randomly partitioned the dataset into 5 groups, each containing 269 images. Subsequently, we used the k-fold method (Olague et al., 2019) with a value of $k = 5$. Our model was trained until convergence for each fold using this dataset. The results obtained from k-fold cross validation are shown in **Tab. 9**. It is evident that the "MAU + MFN-Mamba + Loss" model exhibits minimal variation in IoU and Dice across different folds, indicating consistent performance across diverse data partitions. This observation suggests that the model exhibits robust generalization capabilities and performs effectively on unseen data.

The feature map in the weld cross-sectional reconstruction network is depicted in **Fig. 24**, providing a clear visualization of how this network transforms spatiotemporal feature maps into cross-sectional morphology maps of the weld. In order to further assess the impact of incorporating the new loss function on model performance, this study calculates and compares the values of penetration depth and weld bead area under different process parameters, as shown in **Fig. 25**. The data within the diagram has been normalized based on the maximum value. The $X$ and $Y$ axes represent predicted results and ground truth values respectively; circles denote penetration depth, while triangles mark weld bead shape. It can be observed from **Fig. 25** that the data is uniformly dispersed in close proximity to the line $Y = X$, indicating a negligible discrepancy between predicted values and ground truth. The coefficient of determination (R-Squared accuracy) is presented in the bottom-right corner. The corresponding R-Squared accuracies for penetration depth are determined to be 91.5%, while for weld bead shape it is found to be 92.2%. Upon incorporating the new loss function, there was a notable improvement in R-squared accuracy for both aspects. The R-squared accuracy for penetration depth increased to 95.9%, and for weld bead shape it rose to 94.8%. This indicates that inclusion of the new loss function significantly enhances supervision over model reconstruction in terms of penetration depth.

**Tab. 8** Compared with different methods for weld cross-section morphology reconstruction. Best results are displayed in bold and the runner-up results are underlined.

| No. | Model (feature extraction network + feature fusion network) | $IoU(\%)\uparrow$ | $Dice(\%)\uparrow$ | $HD95\downarrow$ | inference time(ms)↓ |
|---|---|---|---|---|---|
| 1 | CNN-LSTM + MFN-Mamba | 90.05 | 91.38 | 10.84 | **107.42** |

| 2 | ConvLSTM + MFN-Mamba | 91.22(+1.17) | 93.06(+1.68) | 7.62(-3.22) | <u>218.77</u> |
|---|---|---|---|---|---|
| 3 | MAU + MFN-Mamba | <u>94.64(+3.42)</u> | <u>95.43(+2.37)</u> | <u>5.98 (-1.64)</u> | 366.84 |
| 4 | **MAU + MFN-Mamba + Loss** | **96.65(+2.01)** | **96.74(+1.31)** | **4.62(-1.36)** | 360.45 |

**Tab. 9** Results of k-fold cross validation. Best results are displayed in bold and the runner-up results are underlined.

| Model / Time of Fold | CNN-LSTM + MFN-Mamba | | ConvLSTM + MFN-Mamba | | MAU + MFN-Mamba | | MAU + MFN-Mamba + Loss | |
|---|---|---|---|---|---|---|---|---|
| | $IoU(\%)$ | $Dice(\%)$ | $IoU(\%)$ | $Dice(\%)$ | $IoU(\%)$ | $Dice(\%)$ | $IoU(\%)$ | $Dice(\%)$ |
| 1-fold | 90.15 | 91.80 | 91.25 | 93.21 | 94.77 | 95.80 | 95.56 | 96.51 |
| 2-fold | 89.99 | 88.91 | 89.79 | 94.54 | 93.82 | 96.27 | 96.69 | 97.42 |
| 3-fold | 90.98 | 92.36 | 92.02 | 93.88 | 95.01 | 94.19 | 94.89 | 95.99 |
| 4-fold | 90.05 | 90.72 | 90.57 | 92.57 | 94.16 | 96.17 | 95.11 | 96.97 |
| 5-fold | 89.09 | 93.11 | 92.48 | 91.09 | 95.42 | 94.72 | 96.02 | 96.83 |
| Mean | 90.05 | 91.38 | 91.22 | 93.06 | <u>94.64</u> | <u>95.43</u> | **95.65** | **96.74** |
| σ | <u>0.671</u> | 1.633 | 1.083 | 1.323 | **0.645** | 0.926 | 0.723 | **0.533** |

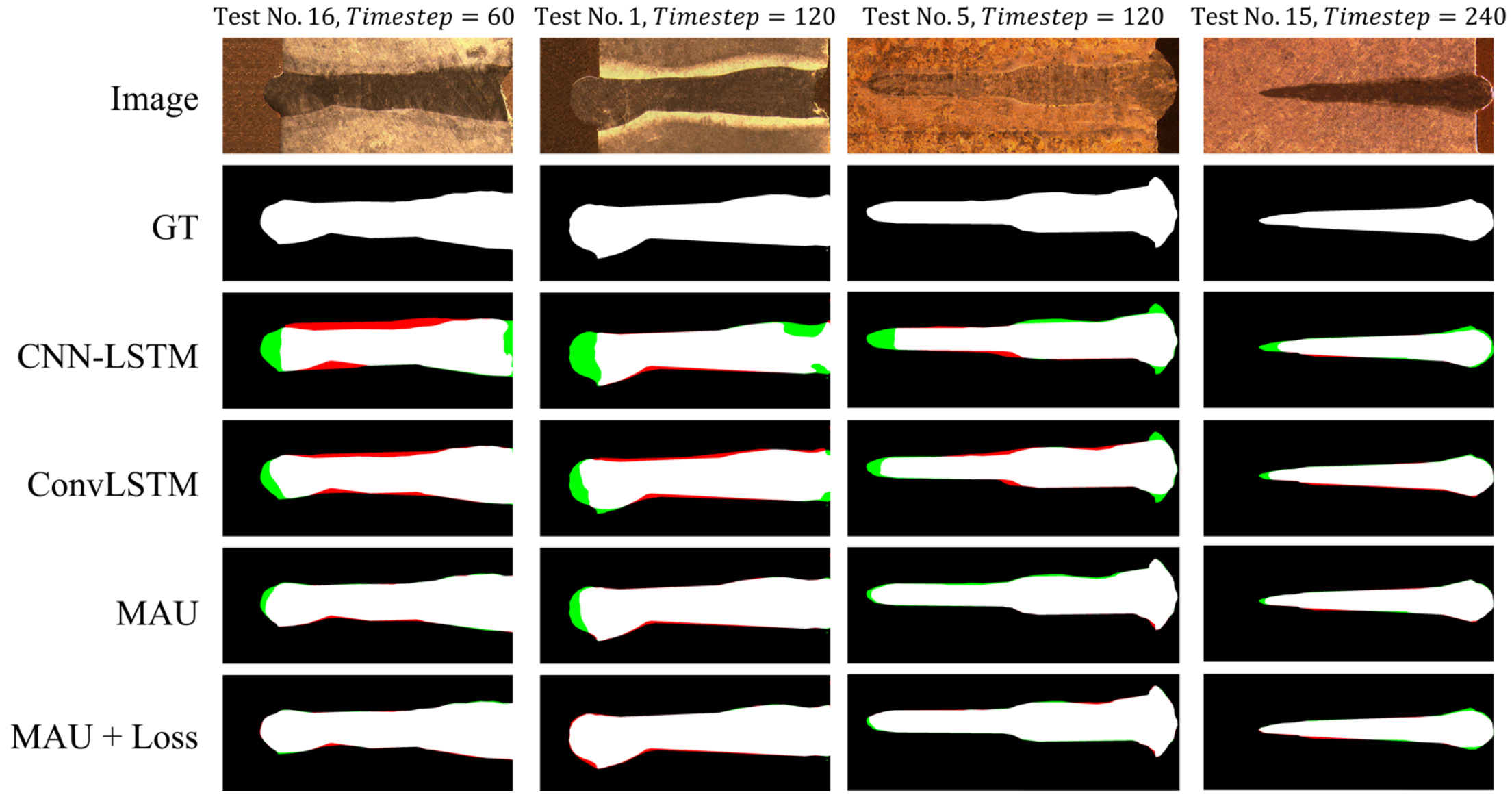

**Fig. 23** Sample results of segmentation on datasets. The colors white, green, and red represent the correct reconstruction, under- reconstruction, and over- reconstruction, respectively.

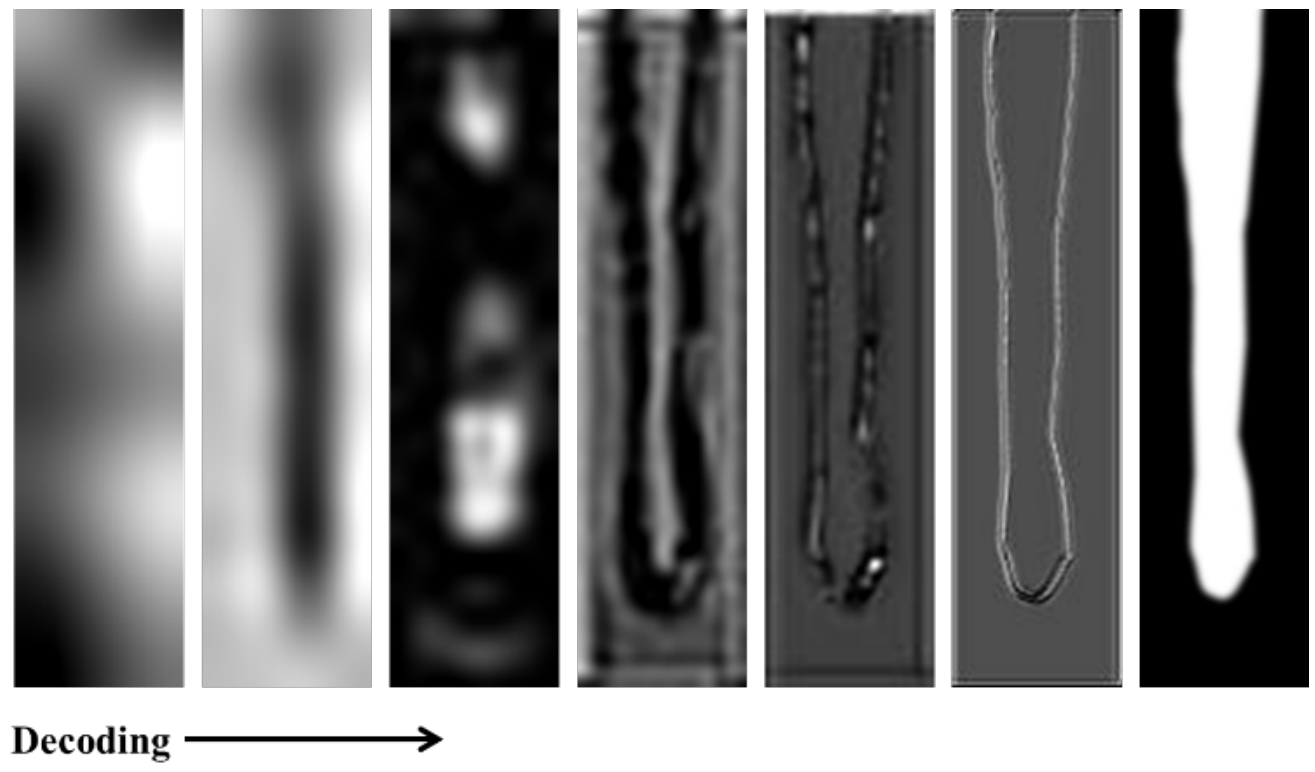

**Fig. 24** Internal decoding process of the weld cross-section reconstruction model under process conditions: $P_d = 9.0kW$; $P_r = 1.0kW$; $V = 1.0m/min$; $D_f = -5.0mm$; $T = 8.0mm, WP = 2$ (training set).

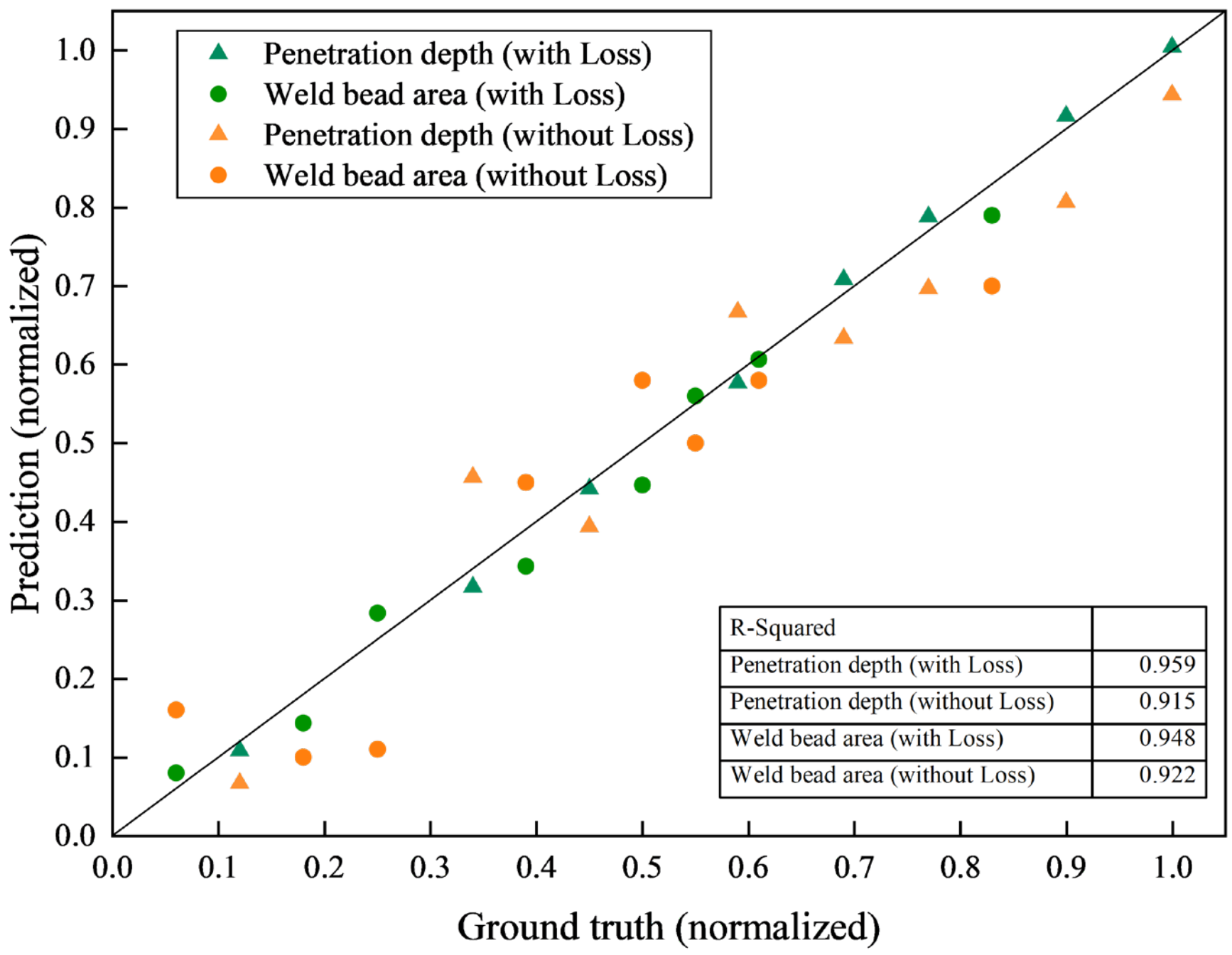

**Fig. 25** Comparison between the prediction result (y-axis) and the ground truth (x-axis) for some test process conditions.

In summary, our model utilizes Mamba exclusively within the feature fusion and cross-sectional reconstruction networks. Future investigations should explore its application in the extraction of features from temporal welding images. Additionally, the study is constrained to predicting the cross-sectional morphology of weld beads through binary images, rather than providing detailed metallographic structures. Subsequent research should consider advanced network architectures, such as Pix2PixHD, to generate high-resolution images with enhanced granular detail. Furthermore, the current study focuses on a single material and a single welding method. A crucial direction for future research involves validating and adapting the proposed model across a broader range of materials and additional welding methods to further establish its generalizability in industrial environments.

## 5. Conclusions

In this study, we utilized a multi-task spatiotemporal deep neural network to extract spatial-temporal features from topside weld pool images and welding parameters. This approach was used for prediction of penetration state, penetration depth, and cross-sectional morphology in laser welding. The welding process, characterized by specific parameters, was treated as an integrated unit, and the captured time-series images were processed into data samples for model training, validation, and testing. The following conclusions can be drawn:

1) An optimal model was proposed for accurately predicting laser welding penetration depth and cross-sectional morphology. The model utilizes efficient spatiotemporal feature extraction and fusion methods to effectively leverage information from weld pool images and welding parameters, achieving stable semantic learning. For the task of predicting penetration status, the accuracy rate is 99.35%, with precision at 99.28%, recall at 99.32%, and F1 score at 99.30%. For the task of predicting penetration depth, the RMSE is 1.79 mm while $R^2$ at 0.9814 respectively. For the task of predicting cross-sectional morphology, the IoU score is recorded as 96.65% while dice score is noted as 96.74%. These results hold great promise for future development of closed-loop control systems in laser welding.

2) In the task of predicting penetration status, the influence of incorporating various welding parameters on model performance was investigated. Notably, laser power had the greatest impact, resulting in a 3.82 increase in accuracy and a 3.92 improvement in F1 score. On the other hand, welding position had minimal effect, with only a marginal enhancement of 0.05 in accuracy and 0.02 in F1 score. Additionally, the impact of adding penetration depth information to the cross-sectional morphology prediction task on accuracy was examined. After validation, both the IoU and Dice scores showed an increase of 1.01% and 1.31%, respectively

**Author contributions** Sen Li: Investigation, Conceptualization, Methodology, Supervision, Writing, Review & editing. Haichao Cui: Investigation, Methodology, Writing, Review& editing. Chendong Shao: Data analysis, Review& editing. Yaqi Wang: Method development, Review& editing. Xihua Tang: Supervision, Review & editing.

**Data availability** The datasets and code from this study can be obtained from the corresponding author upon reasonable request.

## Declarations

**Ethics approval:** Not applicable.

**Consent to participate:** The authors declare their consent to participate.

**Consent for publication:** The authors declare their consent for publication.

**Conflict of interest:** The authors declare no competing interests.

## Funding

The authors gratefully acknowledge the financial support from the National Key R&D Program of China (No. 2023YFB3407800), National Natural Science Foundation of China (No. U2141213).